\documentclass[journal,transmag]{IEEEtran}

\usepackage{xcolor}
\usepackage{color}

\usepackage{flushend}  

\usepackage{cite}

\usepackage[colorlinks,
linkcolor=blue,
anchorcolor=blue,
citecolor=blue]{hyperref}

\usepackage{amsmath}

\usepackage{algorithmic}

\usepackage[ruled,linesnumbered]{algorithm2e}

\usepackage{array}

\usepackage{stfloats}

\usepackage{graphicx} 
\usepackage{float}  
\usepackage{subfigure} 
\graphicspath{ {./images/} }

\usepackage{enumitem}

\usepackage{booktabs}

\usepackage{bbding}
\usepackage{pifont}
\usepackage{wasysym}
\usepackage{amssymb}

\usepackage{multirow}

\usepackage{xr}

\begin{document}
	\title{PixelGame: Infrared small target segmentation as a Nash equilibrium}
	
	\author{Heng Zhou,~\IEEEmembership{Graduate Student Member,~IEEE,}
		Chunna Tian$^*$,
		Zhenxi Zhang,~\IEEEmembership{Graduate Student Member,~IEEE,}
		\\
		Chengyang Li,~\IEEEmembership{Graduate Student Member,~IEEE,}
		Yongqiang Xie$^*$,
		Zhongbo Li
		\thanks{Manuscript received October 14, 2021; 
			revised May 1, 2022.
			\textit{(Corresponding author:Chunna Tian, Yongqiang Xie.)}
		}
		
	}
		\markboth{Journal of \LaTeX\ Class Files,~Vol.~14, No.~8, August~2021}%
	{Shell \MakeLowercase{\textit{et al.}}: A Sample Article Using IEEEtran.cls for IEEE Journals}
	
	
	\maketitle
	
	\begin{abstract}
		A key challenge of infrared small target segmentation (ISTS) is to balance false negative {pixels} (FNs) and false positive {pixels} (FPs). 
		Traditional methods combine FNs and FPs into a single objective by weighted sum, and 
		the optimization process is decided by one actor.
		Minimizing FNs and FPs with the same strategy leads to antagonistic decisions. 
		To address this problem,
		we propose a competitive game framework (pixelGame) from a novel perspective for ISTS.
		In pixelGame, FNs and FPs are controlled by different player whose goal is to minimize their own utility function.
		FNs-player and FPs-player are designed with different strategies: One is to minimize FNs and the other is to minimize FPs.
		The utility function drives the evolution of the two participants in competition.
		We consider the Nash equilibrium of pixelGame as the optimal solution. 
		In addition, we propose maximum information modulation (MIM) to highlight the target information.
		MIM	effectively focuses on the salient region including small targets.
		Extensive experiments on two standard public datasets prove the effectiveness of our method. 
		Compared with other state-of-the-art methods, our method achieves better performance in terms of F1-measure ($\mathrm{F_1}$) and the intersection of union ($\mathrm{IoU}$).
	\end{abstract}
	
	
	\maketitle

	\IEEEdisplaynontitleabstractindextext

	%
	\IEEEpeerreviewmaketitle

	\section{Introduction}
\begin{figure}[t]  
		\centering 
		\subfigure[IR image]{\includegraphics[width=0.21\textwidth]{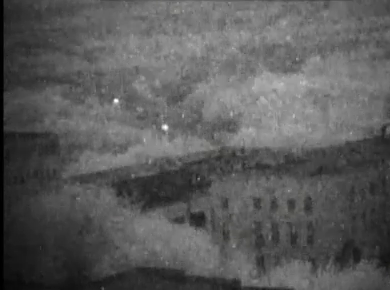}} 
		\subfigure[Ground truth]{\includegraphics[width=0.21\textwidth]{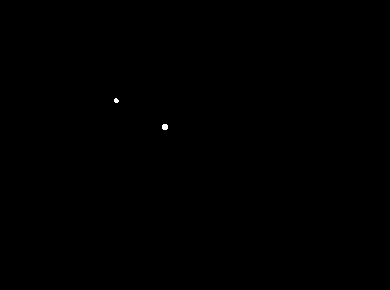}}
		\quad
		\subfigure[Boat RGB image]{\includegraphics[width=0.21\textwidth]{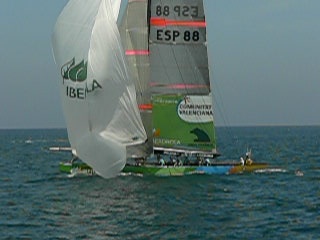}}
		\subfigure[Ground truth]{\includegraphics[width=0.21\textwidth]{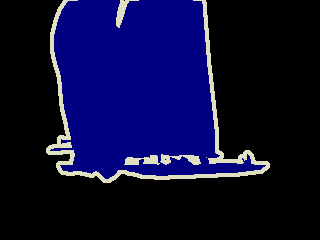}}
		\caption{Comparison between IR imagery from \cite{dai2021asymmetric} and 
			visible boat image from PASCAL VOC 2012 \cite{everingham2011pascal}. 
			The image size is the same in pixel, and on the right is the pixel-by-pixel mask corresponding to the image. 
			Compared with RGB images,
			IR targets are smaller in size, weaker in energy, and sparser in layout.
		} 
		\label{Fig.VOC_IR}  
	\end{figure}

	\IEEEPARstart{I}{nfrared} target segmentation plays a fundamental role in many applications, 
	including surveillance and reconnaissance \cite{razakarivony2016vehicle}, \cite{chen2020survey}, 
	precise strike and guidance \cite{khaledian2014new} in the military field, 
	organ segmentation \cite{yu2018recurrent}, \cite{hesamian2019deep} and
	cell identification \cite{shen2019automatic} in the biomedical field. 
	In real applications, due to the long distance, InfraRed (IR) targets are usually ``dim", ``small" and ``sparse" compared with RGB images.
	The top of Fig. \ref{Fig.VOC_IR} is a typical IR image \cite{dai2021asymmetric}, 
	and the bottom is a common RGB image of natural scene \cite{everingham2011pascal}. 
	{\textit{1) dim}:} 
	The noisy and clutter background results in that the IR targets have low contrast and low signal-to-clutter ratio.
	{\textit{2) small}:} The pixels of IR targets only account for a small proportion in the image. 
	Most of the pixels in IR image are background pixels. 
	{\textit{3)~sparse}:} In addition, it can be seen from Fig. \ref{Fig.VOC_IR} that, different from most RGB images, the layout of IR small targets is also sparse.
	
	Many traditional methods rely on hand-crafted features.
	Traditional methods cannot be adapted to an open and diverse environment due to the lack of texture and shape features.
	They mainly either simplify the small target as a bright spot~\cite{rivest1996detection},~\cite{bai2010analysis},
	or model the background, target and the relationship between them \cite{chen2013local} in a particular scene. 
	Only under specific a priori hypothesis, such methods can achieve good performance. 
	However, in the open environment with the diversity of background scenes, 
	it is difficult to segment IR small targets robustly and accurately.
	
	Different from traditional methods, deep convolutional neural networks (CNN) learn infrared small target representations in a data-driven manner.
	In recent years, inspired by the superior performance of deep learning in machine learning, CNN-based approaches have made new advances in infrared small target detection \cite{liu2020multi} 
	and segmentation \cite{guo2018small}, \cite{fan2018dim}, \cite{hamaguchi2018effective},~\cite{wang2019miss}.
	
	\IEEEpubidadjcol
	
	One of the main challenges is the foreground-background imbalance problem in the ISTS. 
	The foreground pixels in the image are far fewer than the background pixels. 
	Specifically, a large number of background pixels are {incorrectly segmented} as targets (false positive {pixels}, $ FPs $).
	A small number of target pixels are submerged by clutter (false negative {pixels}, $ FNs $).
	
	In order to balance $ FNs $ and $ FPs $, most previous CNN-based methods\cite{ma2021loss} combine the two objectives into one function through weighted sum. 
	The combined objective functions include Dice loss~\cite{isensee2021nnu}, Jaccard loss~\cite{rahman2016optimizing}, 
	Tversky loss~\cite{salehi2017tversky}, asymmetric similarity loss \cite{hashemi2018asymmetric}, 
	sensitivity-specificity loss \cite{brosch2015deep} and penalty loss \cite{yang2019major}.
	The methods based on the combined loss function tune-up the weights to get acceptable solutions, the selection being done by the one actor. 
	The weights as hyperparameters of the loss function control the tradeoff between $ FNs $ and $ FPs $. 
	However, such training objective design mainly suffers from two limitations. 
	1) The same strategy to minimize $ FNs $ and $ FPs $ simultaneously leads to antagonistic decisions.
	The former aims to predict as many target pixels as possible,
	while the latter tends to predict a small number of target pixels with high confidence.
	2)~Extensive studies on loss function show that the setting of their hyper-parameters is an experienced and difficult work. 
	Therefore,
	it is intuitive and rational to optimize $ FNs $ and $ FPs $ as two objectives independently, as in our work.
	
	Our work is motivated by the game theory. 
	Game theory is a framework or paradigm to solve multi-objective optimization problems, 
	especially in dealing with antagonistic criteria \cite{kallel2014nash}, \cite{wang2016multi}, \cite{hsieh2021adaptive}. 
	Game-theoretic learning uses an equilibrium state instead of the optimal solution.
	Following the above idea, we design two tailored sub-networks to act as the FNs-player and FPs-player in the game.
	FNs-player and FPs-player focus on false negative pixels and false positive pixels, respectively. 
	The players, actions and utility functions in the game theory are corresponding to sub-networks, the change of network parameters and loss functions in the proposed framework, respectively \cite{tembine2019deep}, \cite{gemp2021eigengame}. 
	In this way, the ISTS is transformed into a game paradigm.
	Under the constraints of the utility function, game players choose the action that minimizes their utility. 
	Finally, the opponents reach the Nash equilibrium, which is an ingenious tradeoff between $ FNs $ and $ FPs $.
	
	\begin{figure}[b]  
		\centering 
		{\includegraphics[scale=0.13]{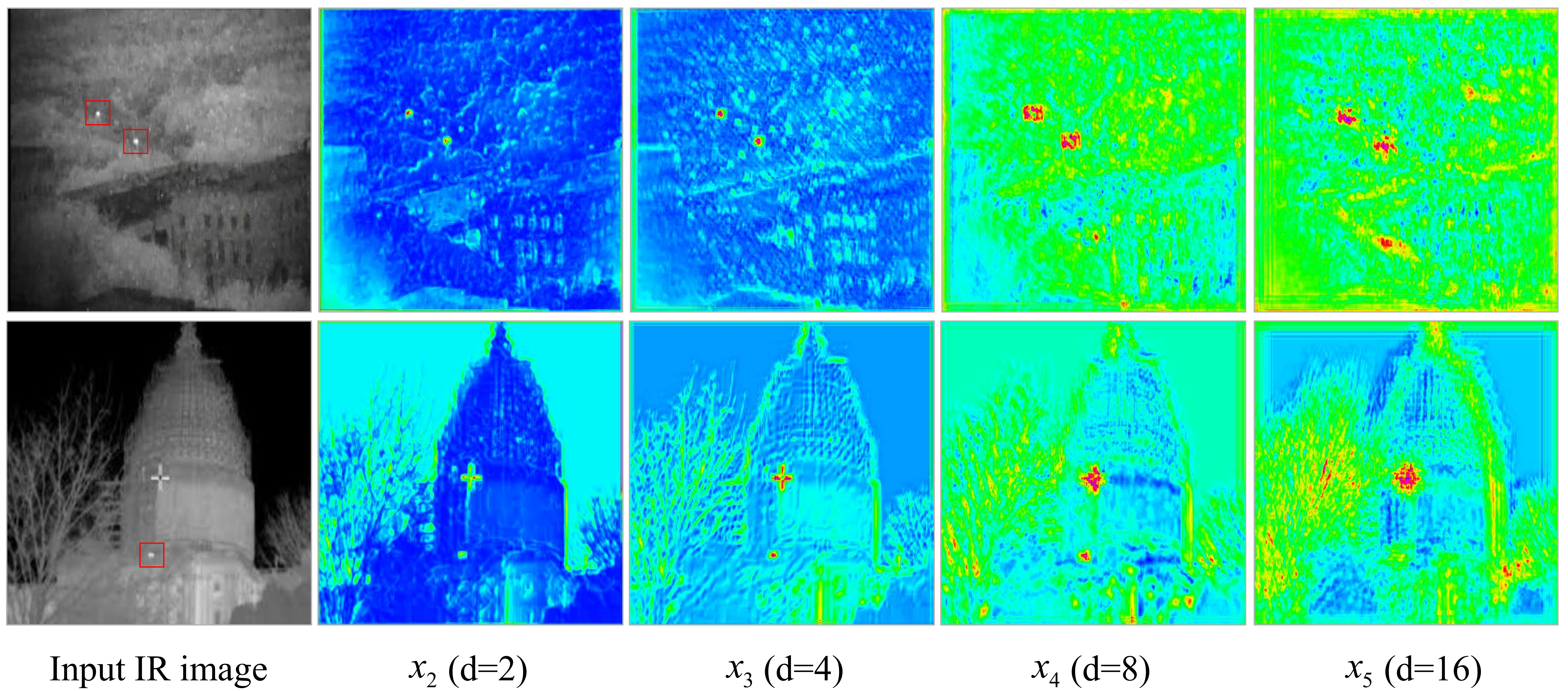}}
		\caption{{
				\textbf{Visualization of feature maps with different dilation factors from dilated convolution.}
				The visualized feature maps with the same resolution are generated by the most responsive channel.
				$ d $ denotes dilation factor.
				From left to right are input images, low-level features ($ x_2 $), middle-level features ($ x_3 $) and high-level features ($ x_4 $ and $ x_5 $) of $\mathrm{FDCN_{9}}\{x_{1}, x_{2}, ..., x_{5}, z_{4}, ..., z_{1}, o \}$, where $ x_n $ and $ z_n $ are the feature map of each layer of the encoder and decoder, respectively.
				The infrared dim small targets are marked with red boxes in input images.
				This figure shows that with the receptive field of dilated convolution increases, the high-resolution features effectively retain the target features.}
		}
		\label{Fig.Visualization}
	\end{figure}
	
	To obtain high-quality segmentation masks, the context of image is important for small targets \cite{hu2017finding}, \cite{mittal2019semi}, \cite{zhao2019m2det}. 
	In deep CNN, the receptive field is mainly expanded by down-sampling. 
	Though, traditional deep convolutional networks have a larger receptive field to aggregate contextual information, they lose some spatial location information. 
	Different from large-scale targets, 
	resolution degeneration may cause that IR small targets fail to be segmented.
	{
	To solve the contradiction between deep high-level semantics and shallow high-resolution feature maps, 
	we adopt the dilated convolution modules \cite{hamaguchi2018effective}, \cite{zhang2017dilated} in both FNs-player and FPs-player. 
	Fully dilated convolution network ($ \mathrm{FDCN} $) can obtain larger receptive fields and maintain the spatial resolution at the same time.
	Fig. \ref{Fig.Visualization} illustrates the visualization of the dilated convolutional feature maps from $ \mathrm{FDCN} $.
	As shown in Fig.~\ref{Fig.Visualization}, as the receptive field of dilated convolution increases, the small targets are effectively retained and enhanced in the high-resolution features.
	}
	
	In addition, in contrast to the obvious semantic dependencies between objects and backgrounds (e.g., boats and rivers, cars and roads, etc.) in RGB object segmentation, 
	the targets in ISTS are grouped into one broad class.
	The semantic contrast between backgrounds and targets is weak in ISTS.
	We observe that small infrared targets are usually local salient in a specific region.
	Based on these observations, we propose maximum information modulation (MIM). 
	MIM absorbs the advantages of attention mechanism \cite{cao2019gcnet}, \cite{misra2021rotate} in focusing on effective information.
	MIM effectively suppresses irrelevant information and enhances the representation of small target in the proposed framework.
	
	In summary, the main contributions of this article are given as follows.
	{
	\begin{enumerate}  
		\item	We present a novel perspective to model infrared small target segmentation as a multi-player strategy game (pixelGame).
		FNs-player and FPs-player focus on reducing $ FNs $ and $ FPs $ in pixelGame, respectively.
		\item	At the same time, 
		a new utility function of pixelGame is designed to encourage two players to conduct the game.
		The utility function ensures that the participants fully play the game and eventually reach Nash equilibrium.
		\item	To handle the small targets, we adopt the dilated convolution modules in both FNs-player and FPs-player. 
		$ \mathrm{FDCN} $ takes both large receptive field and high-resolution feature map into account.
		\item	Due to the fact that IR targets are usually local salient regions in images, 
		we propose MIM to suppress irrelevant background information by calculating local maxima,
		which improves the feature discrimination ability on small targets.
	\end{enumerate}}
	
	The remainder of this article is organized as follows.
	In Section \ref{section.Related_works}, 
	we briefly review the related work on ISTS , deep learning based game theory and two benchmark datasets.
	Section \ref{section.Implementation_details} gives a meticulous description of proposed pixelGame model on ISTS.
	Section \ref{section.Experimental}, 
	we conduct extensive experiments on ablation study and comparison with state-of-the-art (SOTA) methods on the benchmark datasets. The results prove the effectiveness of our method.
	The conclusions and future works are drawn in Section \ref{section.Conclusion}.
	
	\section{Related works}
	\label{section.Related_works}
	In this section, we briefly review related works on ISTS,
	deep learning based game theory and two benchmark infrared dim small target datasets.
	
	\begin{figure*}[b]  
		\centering 
		\includegraphics[scale=0.5]{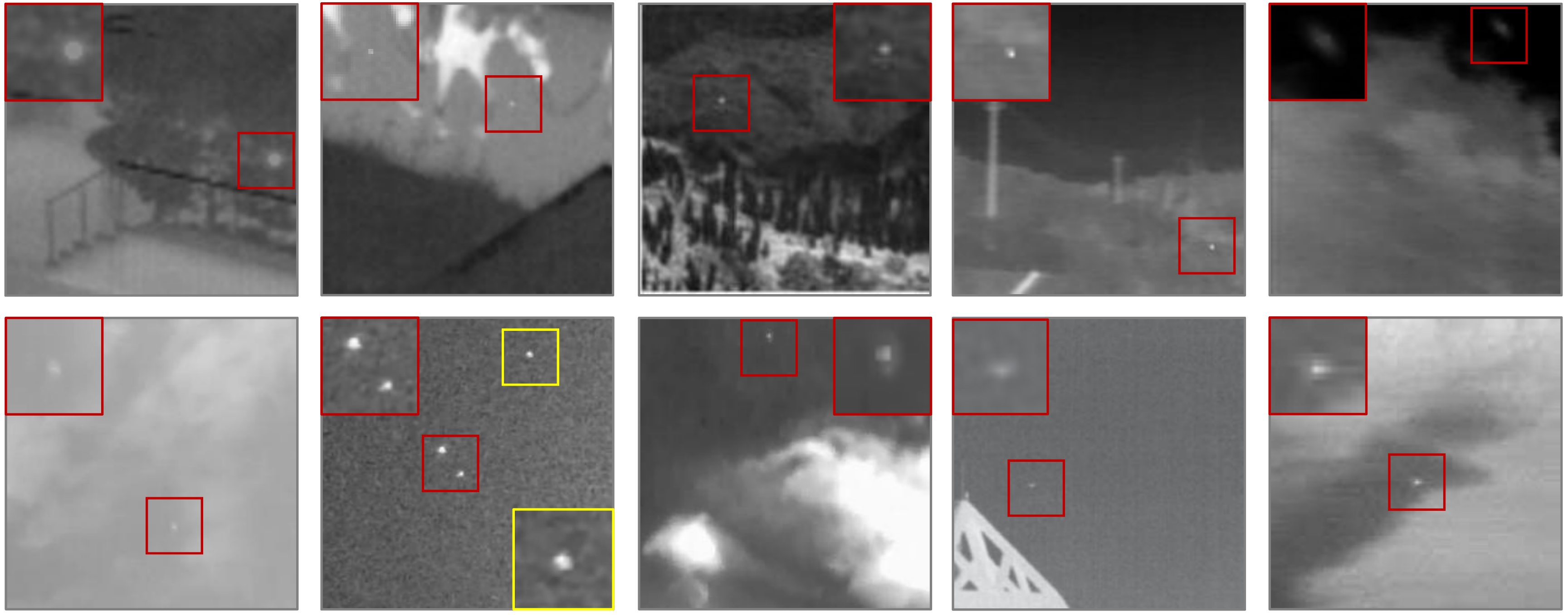}
		\caption{Representative IR images with various backgrounds and targets from the NUST-ISTS \cite{wang2019miss} and NUAA-ISTS \cite{dai2021asymmetric} datasets. 
			For better visualization, the framed area is magnified. The top is from NUST-ISTS, and the bottom is from NUAA-ISTS.
		} 
		\label{Fig.datasets_show} 
	\end{figure*}
	
	\subsection{Infrared small target segmentation}
	
	Infrared small target segmentation models are divided into two categories: 
	Traditional methods based on mathematical modeling with strong prior assumptions and CNN-based methods emerging in recent years.
	Specifically, the traditional ISTS methods mainly include spatial domain filtering methods and optimization-based methods.
	
	Traditional spatial domain-based methods, such as Top-Hat filtering \cite{rivest1996detection}, Max-Median filtering \cite{deshpande1999max}, focus on suppressing the background. 
	Compared with the classic Top-Hat filter, 
	\textit{Bai et al.} \cite{bai2010analysis} constructed a ring filter template through two related but different structural elements to better suppress the background and noise. 
	\textit{Deng et al.} \cite{deng2016infrared} weighted the image entropy by multiscale gray difference to improve the SCR of small targets. 
	\textit{Gao et al.} \cite{gao2018infrared} treated the target as a special sparse component in the noise, so as to distinguish the target from similar background noise. 
	\textit{Huang et al.} \cite{huang2019infrared} used a relatively large density gap between targets and their neighbors to eliminate the interference caused by clutters in complex backgrounds. 
	\textit{Moradi et al.} \cite{moradi2020fast} proposed a directional approach to enhance the target area and suppress structural backgrounds. 
	{
		Many methods apply multi-scale technology to suppress the background. 
		\textit{Nie et al.} \cite{nie2018infrared} proposed a multi-scale local homogeneity measure to improve the saliency of small targets.
		\textit{Gao et al.} \cite{gao2018robust} utilized multi-scale gray and variance difference metrics to enhance the feature representation of small target and mitigate background fluctuations, which improves the detection accuracy.
		In cluttered background, \textit{He et~al.}~\cite{he2021multiscale} enhanced targets by exploiting multi-scale differences in intensity distribution changes and gray values.
	}
	
	Inspired by the human visual system, many local contrast measure (LCM) based methods \cite{chen2013local}, \cite{han2014robust} have been explored. 
	\textit{Wei et al.} \cite{wei2016multiscale} presented multiscale patch-based contrast measure to increase the contrast between the target and background.
	\textit{Huang et al.} improved LCM from the aspects of multiscale~\cite{han2018infrared}, target shape \cite{han2019local} and the difference between the target and the background \cite{han2020infrared}. 
	\textit{Lu et al.} \cite{lu2020robust} utilized a division scheme of surrounding area to capture the derivative properties of the target. These methods extract the difference between the target and the background from various aspects, but their performance is limited when the background changes dramatically or the target is hidden in the background.
	
	The optimizing methods based on low-rank matrix recovery theory assume that the raw image is generated by a low-rank subspace, and the small targets are formulated as sparse singularity. 
	The infrared patch-image (IPI) model \cite{gao2013infrared} regarded small target detection as an optimization problem of recovering low-rank and sparse matrices. 
	\textit{Dai et al.} \cite{dai2017non} used the partial sum of singular values instead of the nuclear norm of IPI to constrain the low rank background patch image.
	For various highly complex background scenes, \textit{Wang et al.} \cite{wang2017pcp} combined the total variation regularization term and principal component pursuit (TV-PCP) to comprehensively describe background feature. 
	\textit{Wang et al.} \cite{wang2017infrared} analyzed the multi-subspace structure of heterogeneous background data, and proposed a stable multi-subspace learning method (SMSL) based on the internal structure of actual images to improve the robustness of the model.
	Self-regularized weighted sparse (SRWS) \cite{zhang2021infrared} model mined the potential information in the background, and transformed the small target segmentation into the optimization problem of extracting clutter from multiple subspaces. 
	
	Compared with matrix, tensors have more advantages in handling high-dimensional data \cite{sun2019infrared}. 
	To distinguish real targets from background residuals in heterogeneous scenes, 
	\textit{Zhang et al.}~\cite{zhang2020edge} proposed an edge and corner awareness-based spatial-temporal tensor (ECA-STT) model. 
	\textit{Sun et al.}~\cite{sun2020infrared} extended the properties of multi-subspace to infrared patch-tensor (IPT) structure to better characterize the highly heterogeneous infrared image background. 
	\textit{Kong et al.}~\cite{kong2021infrared} promoted t-SVD to multimodal t-SVD and enhanced the accuracy of background rank representation in the IPT model. 
	The more accurate the assumptions, the better the performance of these existing IPT-based methods. 
	Therefore, the performance of the model based on optimization relies on the constructed data structure and prior assumptions. 
	
	\begin{table*}[b]
		\centering
		\caption{several latest ISTS datasets. 
			BC and LSR represent background clutters and low spatial resolution, respectively.}
		\label{table.datasets}
		\begin{tabular}{@{}ccccccccc@{}}
			\toprule
			\multicolumn{1}{c}{Datasets} 				& \multicolumn{1}{c}{Background}  & \multicolumn{1}{c}{Year} & \multicolumn{1}{l}{Samples (train\&val)} 	& \multicolumn{1}{c}{Samples (test)} & \multicolumn{1}{c}{Image size} 		& \multicolumn{1}{c}{Label Type} 				& \multicolumn{1}{c}{Sequence} & \multicolumn{1}{c}{Synthetic/Real} \\ \midrule
			NUST-ISTS{\cite{wang2019miss}}             	& BC \& LSR                       & 2019                     & 10,000                                    	& 100                                & (101$\sim$442)$\times$(96$\sim$327)   & Pixel                          				& \XSolidBrush                            & Synthetic                                  \\
			NUAA-ISTS{\cite{dai2021asymmetric}}         & BC \& LSR                       & 2021                     & 341                                      	& 86                                 & (135$\sim$456)$\times$(96$\sim$349)   & Pixel/Box                      				& \XSolidBrush                            & Real                                  \\
			DSAT{\cite{hui2020dataset}}                 & BC                              & 2019                     & 16,177                                    	& N/A                                & 256$\times$256                        & Centroid                       				& \CheckmarkBold                  & Real                                  \\ \bottomrule
		\end{tabular}
	\end{table*}
	
	Specific assumptions cannot adapt to the open and diversified background environment. 
	The model based on deep learning has a large capacity and contains a variety of different scenes.
	In recent years, the release of multiple infrared small target datasets has promoted the research of methods based on CNN.
	{
	\textit{Dai et al.} \cite{dai2021asymmetric} proposes the first CNN-based single-frame infrared small target segmentation model, and designs an asymmetric context modulation (ACM) module to fuse high-level semantics and low-level details.
	ALCNet~\cite{dai2021attentional} transforms the traditional local contrast measurement method into a nonlinear module in the convolution network by combining domain knowledge, which alleviates the problem of the minimum internal characteristics of the pure data-driven methods.
	ISTDU-Net \cite{hou2022istdu} improves the U-Net \cite{ronneberger2015u} segmentation model by increasing the response of small target features and suppressing similar background information, which improves the recognition ability to small targets.
	}

	{	
	ACM, ALCNet and ISTDU-Net take ResNet-20 \cite{he2016deep} as the backbone network to extract infrared small target features, and use the IoU-based weighted loss function to guide model optimization.
	Nevertheless, there are obvious differences in size, energy and layout between infrared and visible targets.
}
	
	{	
	On the one hand, the ResNet structure obtains a large receptive field to fuse context information by sacrificing spatial resolution.
	Different from large-scale RGB targets, the resolution degeneration may cause IR small targets fail to be segmented. 
	On the other hand, the IoU-based loss function minimizes $ FNs $ and $ FPs $ simultaneously, which increases the difficulty of model optimization and leads to antagonistic decisions. 
	Minimizing $ FNs $ aims to predict as many target pixels as possible, while minimizing $ FPs $ trends to predict less number of high-confident target pixels.
	}
	
	{
	Under the framework of generative adversarial networks (GAN) \cite{goodfellow2014generative},
	MDvsFA-cGAN \cite{wang2019miss} used two different segmentation networks as generators, and aims to balance the results of the two generators through adversarial learning between the generators and the discriminator. 
	MDvsFA-cGAN tries to find the Nash equilibrium of the generator and discriminator.
	In addition, GAN networks are difficult to train and prone to model collapse, which causes the generators to produce samples in the same mode.
	Differently, our method aims to find the equilibrium state of the two segmentation networks. 
}

	{
	The core idea of GAN is game theory \cite{kallel2014nash}. 
	To take advantage of game theory, we design two novel segmentation networks as two players. 
	Under the guidance of the utility function, the two networks play games directly. 
	When two players reach Nash equilibrium in the game, the final results are generated by the segmentation networks. 
	}
	
	\subsection{Deep learning based game theory}
	Generally, the deep learning based game is simplified,
	which consists of three parts~\cite{tembine2019deep}: 
	1) the participants of the game are neural units or neural networks; 
	2) the choices of each participant; 
	3) the objective function of each participant.
	A game in strategic form is given as follows:
	\begin{equation}
		\left(\mathcal{T},\left(\varTheta_{k}\right)_{k \in \mathcal{T}},\left(\varPhi_{k}\right)_{k \in \mathcal{T}}\right),
	\end{equation}
	where $\mathcal{T}\{1,2, \ldots\}$ is a set of players.  
	$\varTheta_{k}$ is the set of actions of player $k \in \mathcal{T}$, 
	which is essentially a weight.
	$\varPhi_{k}$ is a loss function of the player $k$. The strategy of player $k$ is to optimize~$\varPhi_{k}$.
	
	Inspired by the outstanding performance of game theory on principal component analysis \cite{gemp2021eigengame}, 
	we regard the task of infrared small target segmentation as a competitive game.
	Different networks as game players focus on different incorrectly segmented pixels. Under the guidance of the utility function, the two players continue to play the game, and finally reach the Nash equilibrium to output the segmentation results.
	
	\subsection{Infrared dim small target datasets}
	
	\begin{figure*}[t]  
		\centering 
		\subfigure[Target Number Distribution]{\includegraphics[width=5.4cm]{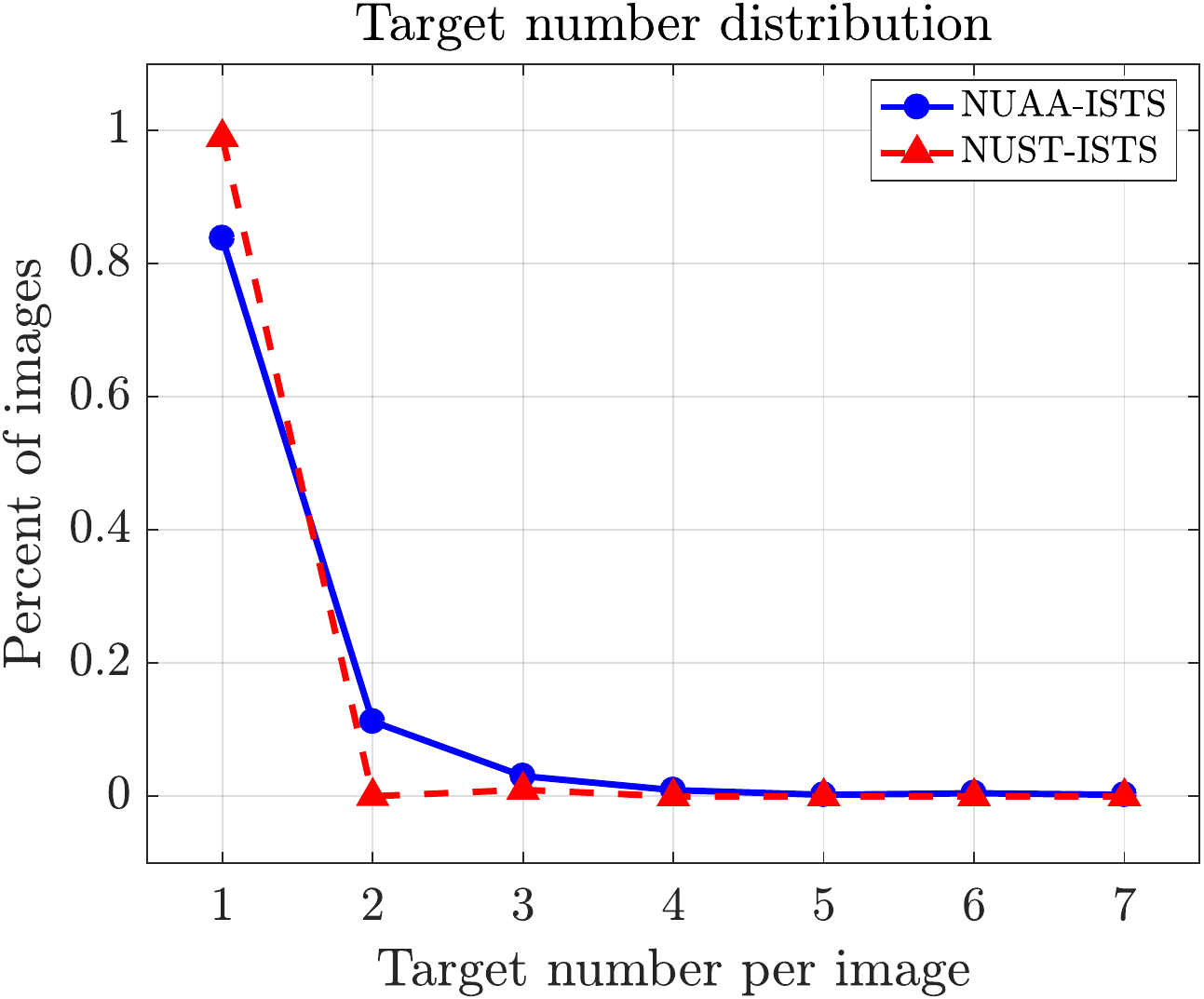}}
		\subfigure[Target Size Distribution]{\includegraphics[width=6.1cm]{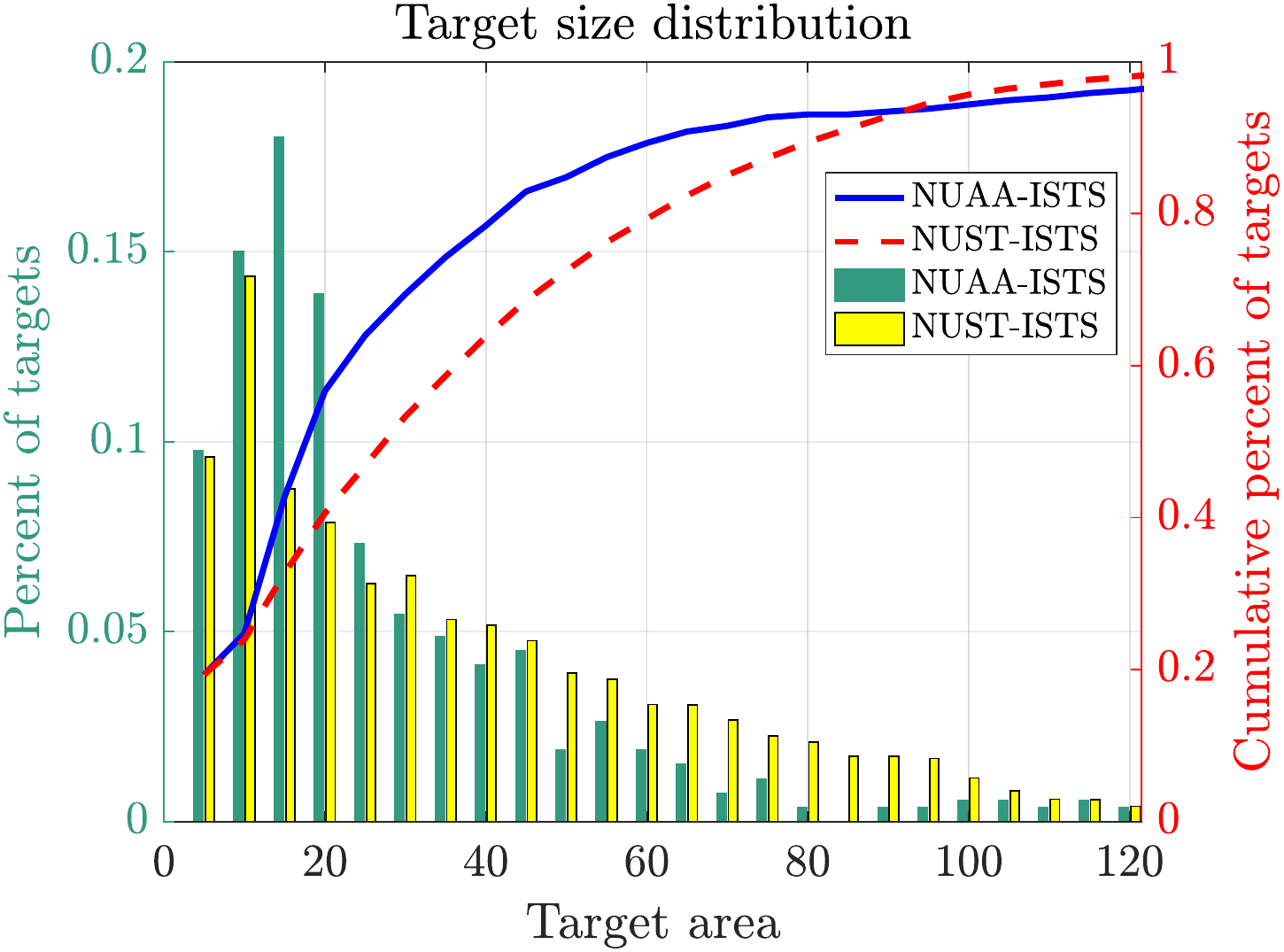}}
		\subfigure[{Target SCR Distribution}]{\includegraphics[width=6.1cm]{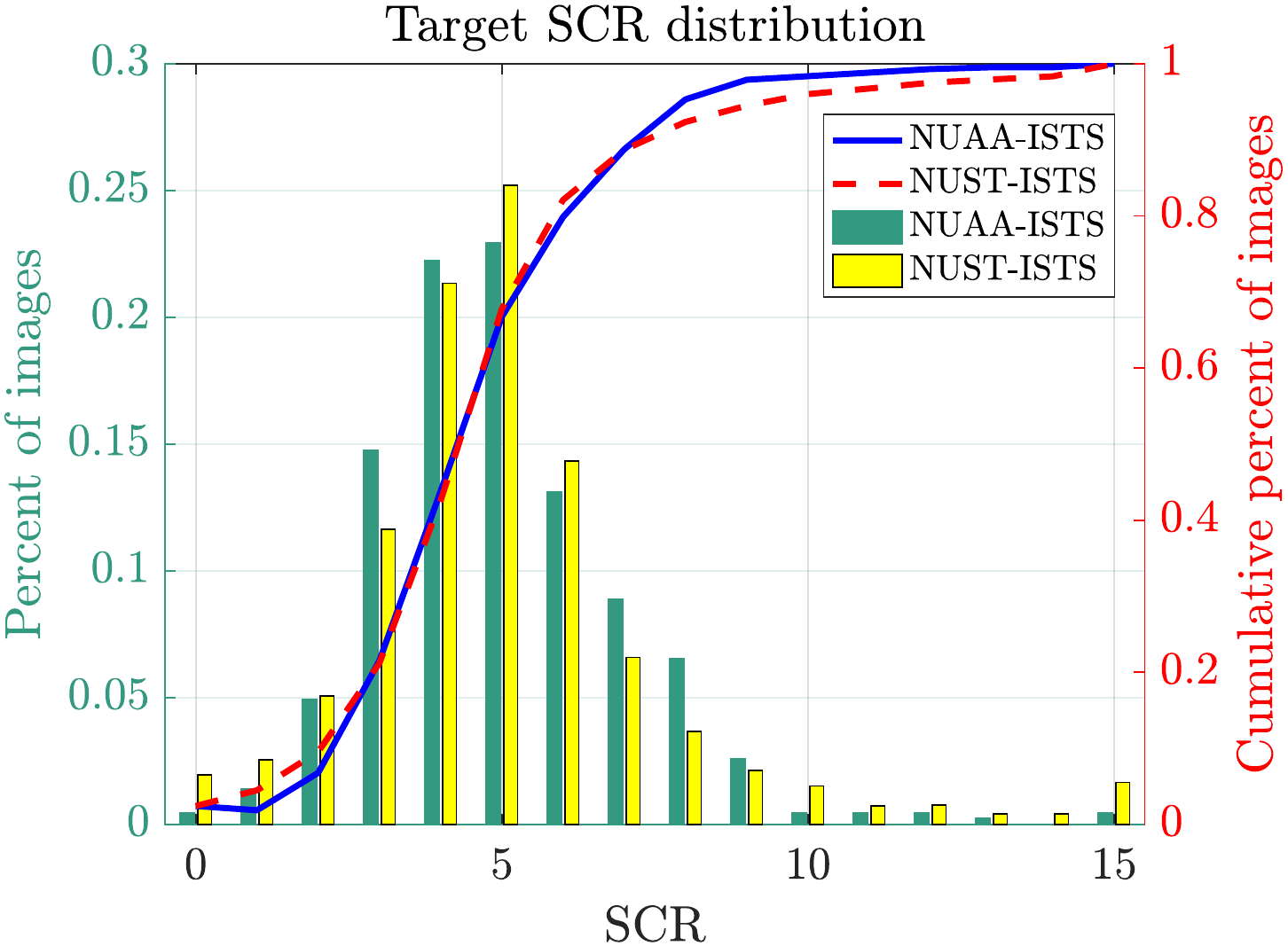}}
		\caption{Illustration of several latest ISTS datasets statistics.
		} 
		\label{Fig.datasets} 
	\end{figure*}
	
	Lacking large-scale data sets severely hampered the application of deep learning on small infrared targets. 
	The recently emerged NUST-ISTS \cite{wang2019miss} is the first large-scale dataset for infrared small target segmentation. 
	NUST-ISTS contains small targets with various real backgrounds, which results in rich data samples.
	\textit{Hu et al.} \cite{hui2020dataset} released the first multi-scene dataset for low slow infrared dim-small aircraft target image sequence (DSAT), the centroid coordinates of objects are marked. 
	NUAA-ISTS \cite{dai2021asymmetric} is the first real scene infrared small target dataset. NUAA-ISTS provides various annotations including mask and bounding box.
	The detailed properties of those three datasets are shown in TABLE \ref{table.datasets}.
	
	In terms of dataset size, the NUST-ISTS and DSAT data sets have more than 10,000 labeled image samples. 
	The difference between them is that the NUST-ISTS dataset is per-pixel labeled, 
	while DSAT only provides the center coordinates of targets. 
	The advantage of the NUAA-ISTS dataset is that the samples cover a variety of natural scenes, 
	and all samples with various sizes are taken from the real background. 	
	
	The experiments of our work are mainly carried out on NUST-ISTS and NUAA-ISTS datasets with pixel-wise target masks. 
	Some representative infrared images of the two datasets are shown in Fig. \ref{Fig.datasets_show}. 
	
	To further analyze the two datasets, we statistically count the number and size of the targets.
	The distribution of the number of targets in each image in those two datasets is shown in Fig.~\ref{Fig.datasets} (a). 
	It can be observed that more than 80\% images contain only one target.
	In detail, the images in NUST-ISTS mainly contain single target, while images in NUAA-ISTS usually contain multiple targets. 
	Fig. \ref{Fig.datasets} (b) shows the statistical distribution of the target area (number of pixels contained in the target) on each dataset.
	From the cumulative line chart on the right, 
	we found that more than 90\% infrared targets contain less than 100 pixels, 
	which occupy less than 1\% in the image. 
	We can see that the targets of infrared images are small, dim and scattered. 
	It can be seen from Fig. \ref{Fig.datasets} (b) that more than half of the targets contains about 20 pixels. 
	
	{The signal-to-clutter ratio (SCR) \cite{gao2013infrared}, \cite{deng2017entropy} is used to measure the target intensity and background intensity.
		In general, the higher SCR of the target, the easier the target is to be segmented.}
	{
	In infrared dim small target segmentation, SCR is defined as follows,
	\begin{equation}
		\textrm{SCR}=\frac{\left|\mu_{t}-\mu_{c}\right|}{\sigma_{c}}.
		\label{Eq.SCR}
	\end{equation}
	In Eq. \eqref{Eq.SCR}, $ \mu_{t} $ represents the target intensity, which is the mean gray value of the target region. 
	$ \mu_{c} $ and $ \sigma_{c} $ represent the mean and standard deviation of the gray value in the target neighborhood region, respectively.
	The target neighborhood region is set to be three times the size of the target region in this paper.
}

{
	As shown in Fig. \ref{Fig.datasets} (c), 
	the SCR of about 70\% IR small targets is lower than 5.
}
	The clutter signals, such as clouds, trees, rocks, ground, etc. account for most energy of the IR images.
	The target can be regarded as the local extreme point in a specific region, but its energy is very weak in the global background.
	
	Given the above analysis, Fig. \ref{Fig.datasets} indicates that the challenges of ISTS not only lie in the limited information from target,
	but also lie in the complex and changeful background.
	
	
	\begin{figure*}[t]  
		\centering 
		\includegraphics[width=0.8\textwidth]{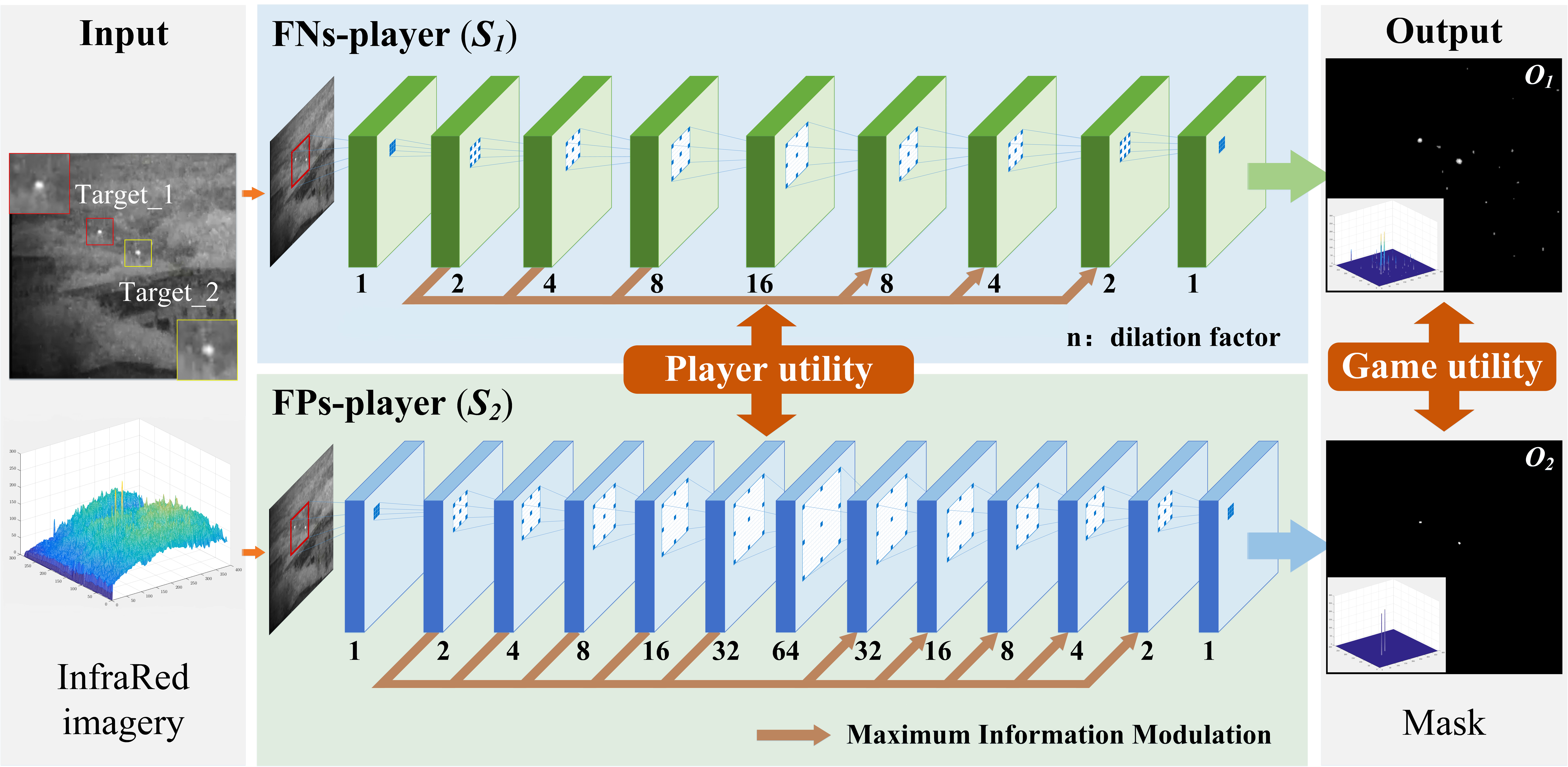}
		\caption{{\textbf{An illustration of the proposed pixelGame.}
				It consists of a FNs-player, a FPs-player and a maximum information modulation.
				The numbers under the feature maps represent the dilation factor of each layer.
				The input and output of the game are IR image and its predicted mask, respectively. 
				In pixelGame, FNs-player and FPs-player are players in cooperative and competitive games. 
				The utility function of pixelGame consists of three parts, including player utility, game utility and small target constraint.
				The player utility guides two players to focus on different pixels, 
				and game utility further enhances the confrontation and competition between the two players.
				In addition, we impose small target constraint on the segmented result by each player.}
		} 
		\label{Fig.pixelGame} 
	\end{figure*}
	
	\section{PixelGame: Infrared dim small target segmentation based on Game Theory}
	\label{section.Implementation_details}
	As shown in Fig. \ref{Fig.pixelGame}, the proposed pixelGame consists of two sub-networks: FNs-player ($ S_1 $) and FPs-player ($ S_2 $), 
	which segment the IR image pixel by pixel under the guidance of respective utility functions. 
	To handle the small targets, we pay attention to deep high-level semantics and shallow high-resolution feature maps at the same time. 
	Therefore, we combine dilated convolution and encode-decode structure to form the backbones of FNs-player and FPs-player.	
	
	In this section, 
	we first introduce how to transform the ISTS task into a game in pixelGame.
	Specially, we deal with three challenges: 
	1) 
	How to design suitable sub-networks to control the focus of different players
	(Section \ref{section.playerNetwork}).
	2)~How to improve the feature representation ability on small targets (Section \ref{section.MIM}). 
	3) How to set a scientific and effective utility function for players to achieve the Nash equilibrium in competitive game (Section \ref{section.utilityfunction}).
		
	\subsection{FNs-player and FPs-player}
	\label{section.playerNetwork}
	Inspired by the game theory in solving antagonistic decisions, 
	two segmentation networks as two players in the game optimize their utility functions, respectively. 
	The sub-networks $S_{1}$ and $S_{2}$ segment the infrared image $\mathbf{I}$ pixel by pixel, as shown in Fig. \ref{Fig.pixelGame}.
	Formally, it can be represented as
	
	\begin{equation}
		\label{Eq.seg}
		\left\{\begin{array}{l}
			S_{1}(\mathbf{I}) \rightarrow \textit{\textbf{O}}_{1}, \\
			S_{2}(\mathbf{I}) \rightarrow \textit{\textbf{O}}_{2},
		\end{array}\right.
	\end{equation}
	where $\textit{\textbf{O}}_{1}$ and $\textit{\textbf{O}}{_2}$ denote to the segmentation results of two players, respectively.
	\begin{table}[t]
		\centering
		\caption{
			Detailed backbone of the networks. In the table, “conv-k(m)-d(n)-c(t)” represents a convolutional layer with m$\times$m kernel, 
			dilation factor of n and output channel number of feature maps of t.
			Head represents the last output layer of the network.
		}
		\label{table.pixelGame}
		\setlength{\tabcolsep}{4mm}{%
			\begin{tabular}{|c|c|c|}
				\hline
				\multicolumn{1}{|c|}{} 	& \multicolumn{1}{c|}{$ \mathrm{FDCN_{9}} $}		& \multicolumn{1}{c|}{$ \mathrm{FDCN_{13}} $} \\[3pt] \hline
				Encoder-decoder
				& \begin{tabular}[c]{@{}l@{}}
					conv-k3-d1-c128  \\
					conv-k3-d2-c128  \\
					conv-k3-d4-c128  \\
					conv-k3-d8-c128  \\
					conv-k3-d16-c128 \\
					conv-k3-d8-c128  \\
					conv-k3-d4-c128  \\
					conv-k3-d2-c128  \\
					conv-k3-d1-c128
				\end{tabular}
				& \begin{tabular}[c]{@{}l@{}}
					conv-k3-d1-c64  \\
					conv-k3-d2-c64  \\
					conv-k3-d4-c64  \\
					conv-k3-d8-c64  \\
					conv-k3-d16-c64 \\
					conv-k3-d32-c64 \\
					conv-k3-d64-c64 \\
					conv-k3-d32-c64 \\
					conv-k3-d16-c64 \\
					conv-k3-d8-c64  \\
					conv-k3-d4-c64  \\
					conv-k3-d2-c64  \\
					conv-k3-d1-c64  \\
				\end{tabular}
				\\ \hline
				Head & conv-k1-d1-c1  &  conv-k1-d1-c1  \\[3pt] \hline
		\end{tabular}}
	\end{table}	
	
	In the small target segmentation task, the $ FNs $ and $ FPs $ are difficult to balance delicately.
	We separate $ FNs $ and $ FPs $, and employ two players to divide and conquer.
	In order to obtain better performance, the two sub-networks use different structures according to tasks. 
	FNs-player and FPs-player use fully dilated convolution network ($ \mathrm{FDCN} $) with different network depth and dilation factor.
	
	The detailed encoder-decoder structures of $ \mathrm{FDCN_{9}} $ and $ \mathrm{FDCN_{13}} $ are shown in TABLE \ref{table.pixelGame}. 
	$ \mathrm{FDCN_{9}} $ and $ \mathrm{FDCN_{13}} $ are the backbone networks of FNs-player and FPs-player, respectively.
	In all of the models, the convolutional layers except the last one are followed by batch normalization (BN)~\cite{ioffe2015batch} and leaky rectified linear unit (leakyReLU)~\cite{zhang2017dilated}.
	Specifically, the goal of $S_{1}$ player is to reduce the false negative pixels of targets, optimizing $ TNs $ and $ FNs $.
	We employ the shallow encoder-decoder network to extract local information and segment all the pixels of the suspected target. 
	The $ \mathrm{FDCN_{9}} $ uses 9-layer convolution, and the dilation factor is increasing from 1 to 16. 
	
	Compared with FNs-player, FPs-player increases the accuracy of the predicted pixels belonging to the target class, by optimizing $ TPs $ and $ FPs $. 
	The pixels predicted by $S_{2}$ may be as precise as possible.
	FPs-player needs a larger context and better local receptive field, so $ \mathrm{FDCN_{13}} $ is deeper and its dilation factor is larger.
	The $ \mathrm{FDCN_{13}} $ contains 13 convolutional layers, and the maximum dilation factor is 64.
	Finally, the head layer is used to predict the class of each pixel, generating the binary mask of foreground and background.
	
	
	\subsection{Maximum Information Modulation}
	\label{section.MIM}
	{
	The information modulation methods represented by the attention mechanism aim to make the model focus on task-related information.
	In RGB object detection and segmentation, 
	SENet \cite{hu2018squeeze} adaptively enhances task-relevant channels by learning the dependencies between different channels.
	Non-local network \cite{wang2018non} is used to capture long-range dependencies and establish the interaction between two pixels with a certain distance on the image.
	GCNet \cite{cao2019gcnet} improves the Non-local network and SENet, enabling query-independent lightweight modules to effectively extract global context information.
	Triplet attention \cite{misra2021rotate} encodes inter-channel relation and spatial relation, and establishes the dependencies between them to calculate attention weights.
	}

	{
		Unlike RGB images, the SCR of infrared dim small targets is very low, and the useful target information is usually submerged in irrelevant clutter and noise. 
		Considering the small targets are arduous to segment, we introduce global max pooling (GMP) \cite{oquab2015object} and 
		cross-channel max pooling ($ \textrm{cMaxPool} $) \cite{misra2021rotate} to enhance the local salient information of these targets.
		The $ \mathrm{MIM} $ aims to increase the pertinence and capacity of extracted features.
		In $ \mathrm{FDCN_{9}} $ and $ \mathrm{FDCN_{13}} $,
		we add the $ \mathrm{MIM} $ module in the skip connection.
	}
	
\begin{figure}[t]  
		\centering 
		\includegraphics[scale=0.13]{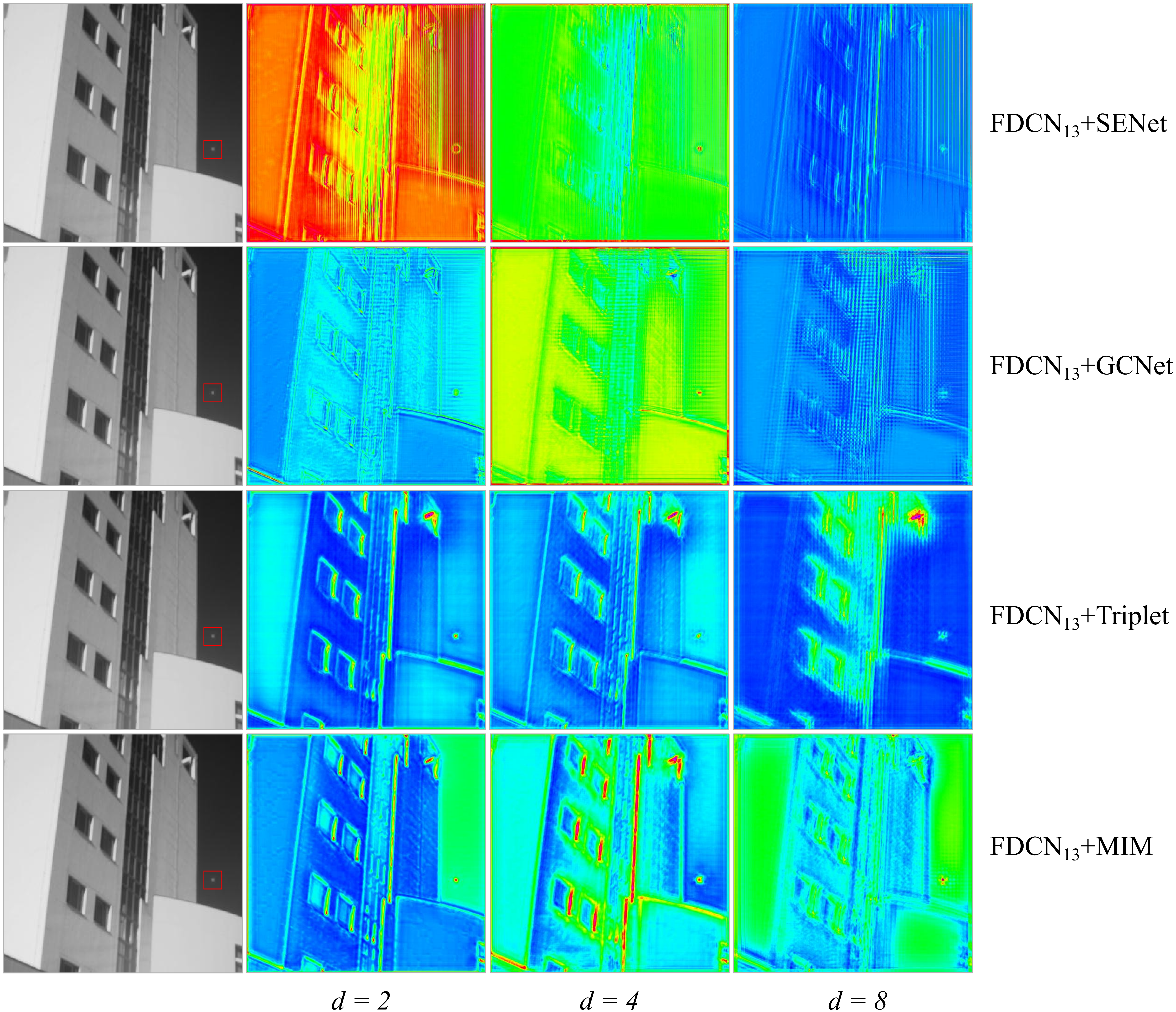}
		\caption{{\textbf{The visualization results of features.}
				We compare the visualization results of MIM module $ \left( \mathrm{FDCN_{13}} + \mathrm{MIM} \right)$ with 
				$ \mathrm{FDCN_{13}} + \mathrm{SENet} $, $ \mathrm{FDCN_{13}} + \mathrm{GCNet} $ and $ \mathrm{FDCN_{13}} + \mathrm{Triplet} $ attention.
				$ d $ denotes the dilation factor of dilated convolution.
				This figure shows that $ \mathrm{MIM} $ prominently enhances the target information and suppresses the background noise in a large number of smooth regions.}
			\vspace{-0.4cm}
		} 
		\label{Fig.Vis_MIM} 
	\end{figure}

	{
		The differences between $\mathrm{MIM}$ and other attention modules are highlighted in Fig. \ref{Fig.Vis_MIM}.	
		It can be seen from the visualization results that the $ \mathrm{MIM} $ performs better in capturing infrared small targets with low SCR than other attention mechanisms.
	}
	
	The $ \mathrm{MIM} $ enhances the salient information related to the target,
	and suppresses a large amount of noise and clutters that are not related to the target.
	{
		In Eq. \eqref{Eq.mim}, the $ \mathrm{MIM} $ is performed on the features $ \mathbf{X} \in \mathbb{R}^{C \times H \times W}$ of each layer of two encoders.
		\begin{equation}
			\mathbf{Z}=m(\mathbf{X}),
			\label{Eq.mim}
		\end{equation}
		where $m(\cdot)$ represent the $ \mathrm{MIM} $, modulation feature $ \mathbf{Z} \in \mathbb{R}^{C \times H \times W}$, $ C, H$ and $ W $ represents the channels, height and width of the feature map, respectively.}
	
\begin{figure}[t]  
		\centering 
		{\includegraphics[width=6.9cm]{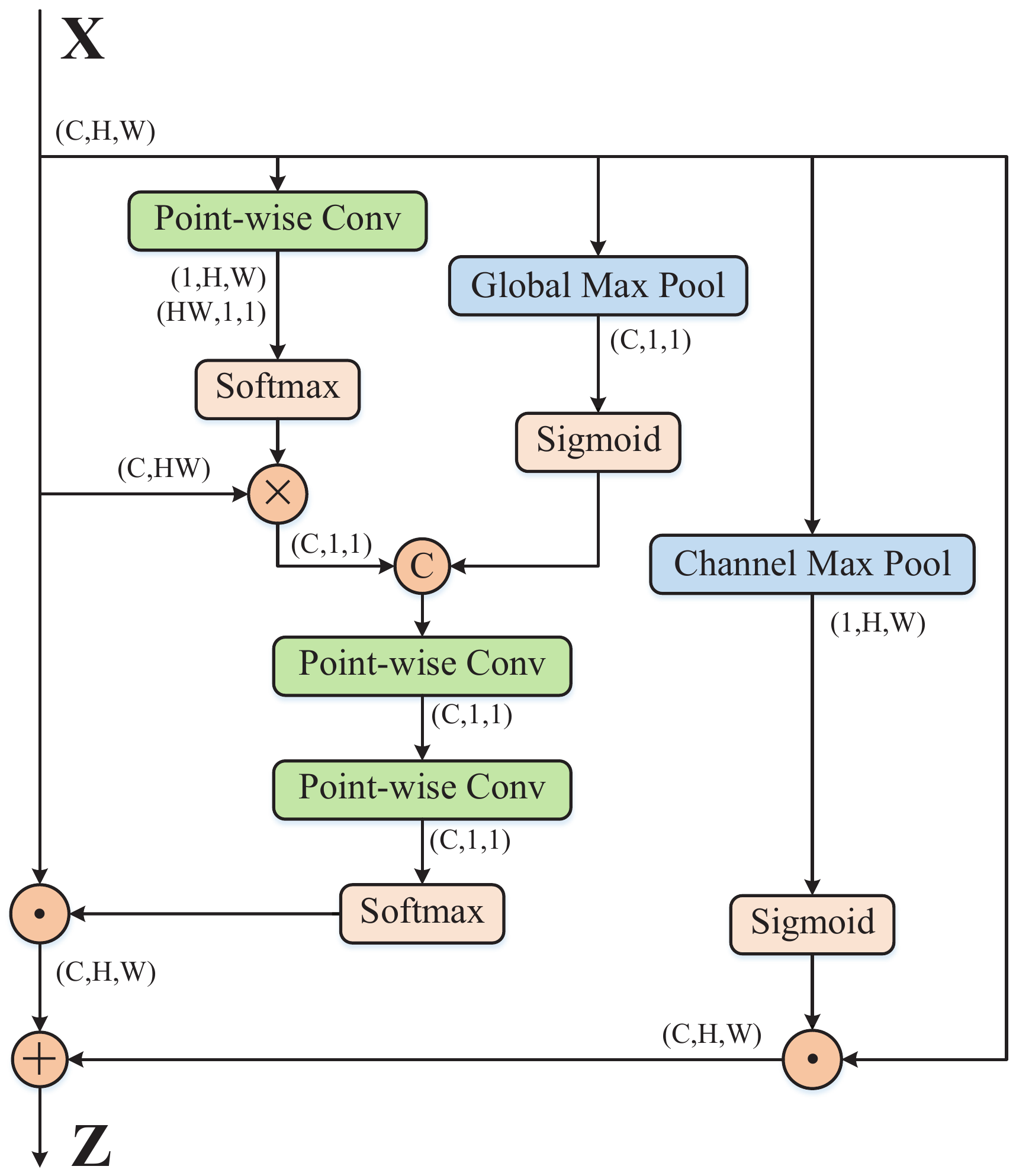}} 
		\caption{{\textbf{The architecture of} $ \mathrm{\textbf{MIM}} $. 
				It combines different attention mechanisms to enhance the salient information of infrared dim small targets.
				$ \otimes $ denotes matrix multiplication,
				$ \copyright $ denotes tensor concatenation,
				$ \odot $ denotes broadcast element-wise multiplication, 
				and $ \oplus $ denotes matrix addition.}
		} 
		\label{Fig.MIM} 
	\end{figure}

	
	The network structure of $ \mathrm{MIM} $ is shown in Fig. \ref{Fig.MIM}.
	First, the pixel-wise correlation in the spatial domain is used to obtain cross-channel attention $\mathbf{V_1} \in \mathbb{R}^{C \times 1 \times 1}$.
	Specifically, in Eq.~\eqref{Eq.PWConv},
	the feature map $\mathbf{X}$ is first transformed through point-wise convolution (PWConv)~\cite{lin2013network} to fuse the relationship between the features of different channels.
	Eq. \eqref{Eq.Reshape} makes the $ \mathrm{MIM} $ pay more attention to the correlation between different spatial positions of feature.
	\begin{equation}
		\mathbf{Y}=\textrm{PWConv}(\mathbf{X}),
		\label{Eq.PWConv}
		\vspace{-0.5cm}
	\end{equation}
	\begin{equation}
		\mathbf{V_1}={reshape}(\mathbf{X}) * \rho\left({reshape}\left(\mathbf{Y}\right)\right),
		\label{Eq.Reshape}
	\end{equation}
	where the PWConv contains BN, and leakyReLU sequences. $\rho$ is the Softmax function. 
	Dimension reshaped as $\mathbf{X} \in \mathbb{R}^{C \times H \times W} \rightarrow \mathbb{R}^{C \times HW}$,
	$\mathbf{Y} \in \mathbb{R}^{1 \times H \times W} \rightarrow \mathbb{R}^{HW \times 1 \times 1}$.
	
	
	{
	Secondly, 
	different from RGB target segmentation, the targets in ISTS are often relatively small.
	GMP selects the extreme values in the feature map, 
	it can effectively improve the saliency of the feature map in the region and suppress noise.
	Global average pooling (GAP)~\cite{zhou2016learning} usually pays more attention to large objects. 
	GAP tends to give higher responses to large irrelevant objects and neglect the small extreme regions. 
	However, the small targets to-be-segmented in our task are usually local extreme points in images, 
	GMP extracts the target-related features more effectively. 
	Thus, GMP can reduce the impact of useless background information and highlight striking target information.
	}
%
	In Eq.~\eqref{Eq.GMP}, we obtain channel-wise attention $\mathbf{V_2} \in \mathbb{R}^{C \times 1 \times 1}$:
	\begin{equation}
		\mathbf{V_2}=\sigma(\textrm{GMP}(\mathbf{X})),
		\label{Eq.GMP}
	\end{equation}
	where $\sigma$ is the Sigmoid function.

	We combine different target information obtained by Eq.~\eqref{Eq.Reshape} and Eq. \eqref{Eq.GMP} 
	to obtain dual-channel attention $\mathbf{M}_\mathbf{1} \in \mathbb{R}^{2C \times 1 \times 1}$:
	\begin{equation}
		\mathbf{M}_{\textbf{1}}=[\mathbf{V_1}; \mathbf{V_2}].
	\end{equation}
	
	As shown in Fig. \ref{Fig.MIM}, the two-layer PWConv realizes the full interaction of different channel information through squeezing and excitation, and further enhances the cross-channel global attention $ \mathbf{M}_{\mathbf{2}} $, in Eq. \eqref{Eq.M2}.
	\begin{equation}
		\mathbf{M}_{\mathbf{2}}=\rho\left(\textrm{PWConv}\left(\textrm{PWConv}\left(\mathbf{M}_{\textbf{1}}\right)\right)\right),
		\label{Eq.M2}
	\end{equation}
	where $\mathbf{M}_\mathbf{2} \in \mathbb{R}^{C \times 1 \times 1}$, and the last layer does not include leakyRelu activation.
	
	Then, we use $ \textrm{cMaxPool} $ to obtain the most significant information of different channels.
	$\mathbf{M}_\mathbf{3}$ in Eq. \eqref{Eq.M3} is the max channel pooling feature map, 
	where each point in the feature map is the max of the points at the same position within feature maps.
	\begin{equation}
		\mathbf{M}_{\mathbf{3}}=\sigma\left(\textrm{cMaxPool}\left(\mathbf{X}, 0\right)\right),
		\label{Eq.M3}
	\end{equation}
	where $\sigma$ is the Sigmoid function, $\mathbf{M}_\mathbf{3} \in \mathbb{R}^{1 \times H \times W}$.
	
	Finally, guided by Eq. \eqref{Eq.M2} and Eq. \eqref{Eq.M3}, $ \mathrm{MIM} $ enhances small target features in both spatial and channel dimensions. 
	The final feature is obtained by
	\begin{equation}
		\mathbf{Z}=\mathbf{M}_\mathbf{2} \cdot \mathbf{X} + \mathbf{M}_\mathbf{3} \cdot \mathbf{X},
	\end{equation}
	where $\mathbf{Z} \in \mathbb{R}^{C \times H \times W}$.

	\subsection{Utility function}
	\label{section.utilityfunction}
	When evaluating segmentation results, the predicted results are often divided into four parts: 
	{
	The number of correctly segmented pixels of target ($ TPs $),
	the number of incorrectly segmented pixels of background ($ FPs $),
	the number of correctly segmented pixels of background ($ TNs $),
	the number of incorrectly segmented pixels of target ($ FNs $).
	The target pixels are composed of $ TPs $ and $ FNs $.
	The background pixels contain $ TNs $ and $ FPs $.
	}
	The confusion matrix of ISTS is shown in TABLE \ref{Table.cMat}.
	In the confusion matrix, the columns represent the predicted masks $ \textit{\textbf{O}} $ of the pixelGame, and the row represents the ground truth $ \textit{\textbf{G}} $ of the input images.
	
	\begin{table}[H]
		\centering 
		\caption{{
				Confusion matrix. 
				Among them, 
				“1” and “0” represent the target and background respectively.}}
		\label{Table.cMat} 
		\setlength{\tabcolsep}{1mm}{%
			{
				\begin{tabular}{|cc|cc|}
					\hline
					\multicolumn{2}{|c|}{\multirow{2}{*}{Confusion matrix}}    & \multicolumn{2}{c|}{\textit{\textbf{G}}}            \\ \cline{3-4} 
					\multicolumn{2}{|c|}{}                                     & \multicolumn{1}{c|}{Actually Positive (1)} & Actually Negative (0) \\ \hline
					\multicolumn{1}{|c|}{\multirow{2}{*}{\textit{\textbf{O}}}} & Predicted Positive (1) & \multicolumn{1}{c|}{$ TPs $}                   & $ FPs $                   \\ \cline{2-4} 
					\multicolumn{1}{|c|}{}                   & Predicted Negative (0) & \multicolumn{1}{c|}{$ FNs $}                   & $ TNs $                   \\ \hline
		\end{tabular}}}
	\end{table}

	In order to achieve high-quality results for the FNs-player and FPs-player games,
	we design a novel utility function according to each player's own focus and the overall constraints of the game. 
	This utility function consists of three parts: 
	Player utility, game utility and small target constraints. 
	Scientific and effective utility function help the model reach the equilibrium state in the competitive game.
	
	
	\subsubsection{Player utility}
	
	For player utility, the main goal of FNs-player and FPs-player is to minimize $ FNs $ and $ FPs $. 
	{Combined with the confusion matrix in TABLE \ref{Table.cMat}, the utility functions $ U $ of players are defined as follows:}
	
	\begin{equation}
		U(\textit{\textbf{O}}_1, \textit{\textbf{G}})=\frac{F N s}{T N s+F N s},
		\label{Eq.P_FNs} 
	\end{equation}
	\begin{equation}
		U(\textit{\textbf{O}}_2, \textit{\textbf{G}})=\frac{F P s}{T P s+F P s}.
		\label{Eq.P_FPs}
	\end{equation}
	
	
	The utility function of FNs-player in Eq. $\eqref{Eq.P_FNs}$ focuses on driving the player to predict as many target pixels as possible. 
	On the opposite, in Eq. (\ref{Eq.P_FPs}),
	the utility of FPs-player makes the player distinguish the target and background more finely.

	\subsubsection{Game utility}
	As the two sides of an antagonistic decision, 
	they need to further enhance the complementary differences between them. 
	Therefore, we define the game utility $ G $ as follows:
	{\begin{equation}
			G\left(\textit{\textbf{O}}_{1}, \textit{\textbf{O}}_{2} \mid \textit{\textbf{G}}\right)=
			\left\|\left(\textit{\textbf{O}}_{1}-\textit{\textbf{G}}\right)\left(\textit{\textbf{O}}_{2}-\textit{\textbf{G}}\right)\right\|_{2}.
			\label{Eq.game}
	\end{equation}}

	Eq. (\ref{Eq.game}) makes the incorrectly segmented pixels of FNs-player and FPs-player as different as possible.
	Game utility further aggravates the game confrontation between them. 
	Explicit antagonistic utility constraints can not only enhance the complementarity of their results, 
	but also help the game optimization to reach the equilibrium state.

	\subsubsection{Small target constraints}
	For the specific task of ISTS, 
	we add a small target constraint to ensure that the model is optimized in a reasonable space.
	It is defined as follows:
	\begin{equation}
		A\left(\textit{\textbf{O}}\right)=\frac{1}{N} \sum_{k=1}^{N} o_{k}.
		\label{Eq.area}
	\end{equation}
	
	{
		Finally, we choose equal weights according to many experimental attempts.
		The utility $ \varPhi $ of pixelGame is as follows,
		\begin{equation}
			\varPhi\left(\textit{\textbf{O}}_{1}, \textit{\textbf{O}}_{2} \mid \textit{\textbf{G}}\right)=
			U\left(\textit{\textbf{O}}, \textit{\textbf{G}}\right)+
			G\left(\textit{\textbf{O}}_{1}, \textit{\textbf{O}}_{2} \mid \textit{\textbf{G}}\right)+
			A\left(\textit{\textbf{O}}\right).
			\label{utility}
	\end{equation}}

	
	\subsection{PixelGame Network}
	$\mathrm{FDCN_{9}}$ and $\mathrm{FDCN_{13}}$ are the backbones of pixelGame network. 
	A high-resolution prediction feature map is indispensable for small target segmentation. 
	The dilated convolution captures a larger receptive field without reducing the spatial resolution of the feature. 
	The dilation factors of the decoder are symmetrical to that of the encoder. In the encoder-decoder structure, 
	the feature mapping with the same dilation factor exchanges information across layers through skip connection.
	
	The larger dilation factor results in a larger receptive field. 
	On the one hand, some pixels in the large receptive field are not fully utilized.
	On the other hand, the long-distance dependence of the pixels captured by the large receptive field is not accurate. 
	Therefore, we use $ \mathrm{MIM} $ module to improve small object feature representation.
	
	The speciﬁc implementation is shown in Algorithm \ref{alg.pixelGame}.
	
	\begin{algorithm}[htbp]
		\caption{{PixelGame: ISTS as a Nash Equilibrium}}
		\label{alg.pixelGame}
		\SetKwInput{KwInput}{Input}                
		\SetKwInput{KwOutput}{Output}              
		\DontPrintSemicolon
		
		\KwData{input image $\mathbf{I}$, ground truth \textit{\textbf{G}}, 
			utility function~$\varPhi$, 
			game player networks~$\varTheta\{{S_1, S_2}\}$, 
			maximum information modulation ($ \mathrm{MIM} $)
		} 
		\KwOut{equilibrium state of the game $\textit{\textbf{O}}$
		}
		$S_1 \leftarrow \mathrm{MIM} ( \mathrm{FDCN_9} )$\;
		$S_2 \leftarrow \mathrm{MIM} ( \mathrm{FDCN_{13}} )$\;
		\ForEach{epoch}
		{
			$\textit{\textbf{O}}_1,\textit{\textbf{O}}_2 \leftarrow S_{1}(\mathbf{I}),S_{2}(\mathbf{I})$\;
			$\varPhi \leftarrow$ Eq. $\eqref{utility}$\;
			$S_1, S_2\leftarrow\arg\min \varPhi$
		}
		$\textit{\textbf{O}} \leftarrow mean(\textit{\textbf{O}}_1,\textit{\textbf{O}}_2)$\;
		\KwRet $ \textit{\textbf{O}} $\;
		
		\;
		
		\SetKwFunction{FMain}{MIM}
		\SetKwProg{Fn}{Function}{:}{}
		\Fn{\FMain{$\mathrm{FDCN_{2n+1}}$}}{
			{
				$\mathrm{FDCN_{2n+1}} = \{x_{1}, ..., x_{n}, x_{n+1}, z_{n}, ..., z_{1}, o \}$,
				where $ x_n $ and $ z_n $ are the feature map of each layer of the encoder and decoder, respectively.\;}
			\ForEach{layer $k \in [2,n]$}
			{
				{$z \leftarrow m(x_k) $ using Eq. $\eqref{Eq.mim}$ \;
					$z_{k}^{\ast}\leftarrow z_k + z$ \;}
			}
			\textbf{return} $S \leftarrow \{x_{1}, x_{2}, ..., x_{n}, x_{n+1}, z_{n}^{\ast}, ..., z_{2}^{\ast}, z_{1}, o \}$\;
		}
		%
		
	\end{algorithm}
	
	\section{Experimental results and analysis}
	\label{section.Experimental}
	
	In order to analyze the potential of pixelGame based on game theory, 
	we compare it with the related SOTA methods on NUST-ISTS and NUAA-ISTS datasets.
	Moreover, we conducted ablation experiments to verify the effectiveness of different components in pixelGame.
	In particular, the following questions will be investigated in our experimental evaluation.
	
	{\begin{enumerate}  
		\item	Q1: Our key insight is to transform the antagonistic decision of $ FNs $ and $ FPs $ using the same strategy into a competitive game in which two player networks participate. 
		Based on the game theory, we study how FPs-player and FPs-player achieve a delicate balance 
		under the guidance of utility function with inherent complementarity and optimization antagonism. (Section \ref{section.NashEq}).
		\item	Q2: We further explore the contribution of the composition of utility function 
		to the Nash equilibrium in pixelGame. (Section \ref{section.Q2}).
		\item	Q3: In an antagonistic decision of pixelGame, 
		whether the performance of a multi-player strategy game is better than 
		that of a single integrated objective through weighted sums. (Section \ref{section.Q3}).
		\item	Q4: 
		We compare our pixelGame with other SOTA methods and show the segmentation results. The effectiveness of $ \mathrm{MIM} $ module is proved. 
		(Section \ref{section.SOAT}).
	\end{enumerate}}

	\subsection{Experimental Settings}
	The experiment is conducted on a computer with 3.0 GHz CPU, 128 GB RAM, and four NVIDIA GeForce RTX 3090 GPUs. 
	The pixelGame is implemented by PyTorch. 
	We use Adam as the optimizer.
	The mini-batch size is set as 8. The learning rate is set as $10^{-5}$ for the FNs-player and FPs-player. 
	The weights of the players are initialized using the identity initialization technique. 
	All models are trained from the scratch, and the training epoch of NUST-ISTS is 70 and that of NUAA-ISTS is 400.

	\subsubsection{Datasets}
	Two pixel-level annotated IR small target datasets with diverse scenes and 
	targets are used to verify the performance of the proposed methods, as shown in TABLE \ref{table.datasets}. 
	
	To increase the number of the training samples, NUST-ISTS randomly sampled many patches with the same size from the original images, 
	which added up to 10,000 patches for training under different configurations \cite{wang2019miss}.
	NUAA-ISTS contained 427 representative images and 480 instances of different scenarios from hundreds of real-world videos \cite{dai2021asymmetric}. 
	Since the images of the training dataset released by the author of NUST-ISTS are cropped, 
	the training data composed of cropped images and the testing data are different in distribution. 
	Therefore, we re-divide the dataset, randomly sample 80\% original training data and the testing data to form the new training dataset, and the rest is the new test dataset. 
	Same as the setting in \cite{dai2021attentional}, in order to stack images of different sizes into a batch, 
	the size of each image is resized to 512$\times$512, and randomly cut to 480$\times$480 during training.
	
	\subsubsection{Implementation Details}
	We compare our proposed pixelGame with several SOTA small target detection and segmentation methods: 23 traditional methods and 3 CNN-based methods.
	
	These methods can be categorized into four groups, including 
	i) Traditional spatial domain-based methods (Top-Hat \cite{rivest1996detection}, MNWTH \cite{bai2010analysis}, 
	Max-Median \cite{deshpande1999max}, NWIE \cite{deng2016infrared}, MoG-MRF \cite{gao2018infrared}, DPIR \cite{huang2019infrared}, 
	ADMD \cite{moradi2020fast}); 
	ii) human visual system-based methods (LCM \cite{chen2013local}, ILCM \cite{han2014robust}, MPCM~\cite{wei2016multiscale}, 
	RLCM \cite{han2018infrared}, TLLCM \cite{han2019local}, WSLCM \cite{han2020infrared}, MDWCM~\cite{lu2020robust}); 
	iii) optimization-based methods (
	IPI~\cite{gao2013infrared}, SMSL \cite{wang2017infrared},
	NIPPS~\cite{dai2017non}, TV-PCP \cite{wang2017pcp}, 
	SRWS \cite{zhang2021infrared}, WSNMSTIPT \cite{sun2019infrared}, 
	ECA-STT~\cite{zhang2020edge}, 
	NTFRA \cite{kong2021infrared}); 
	and iv) CNN-based algorithm (MDvsFA-cGAN \cite{wang2019miss}, ACM \cite{dai2021asymmetric}, ALCNet \cite{dai2021attentional}).
	
	\begin{figure*}[t]  
		\centering 
		\subfigure[NUST-ISTS]
		{\includegraphics[scale=0.65]{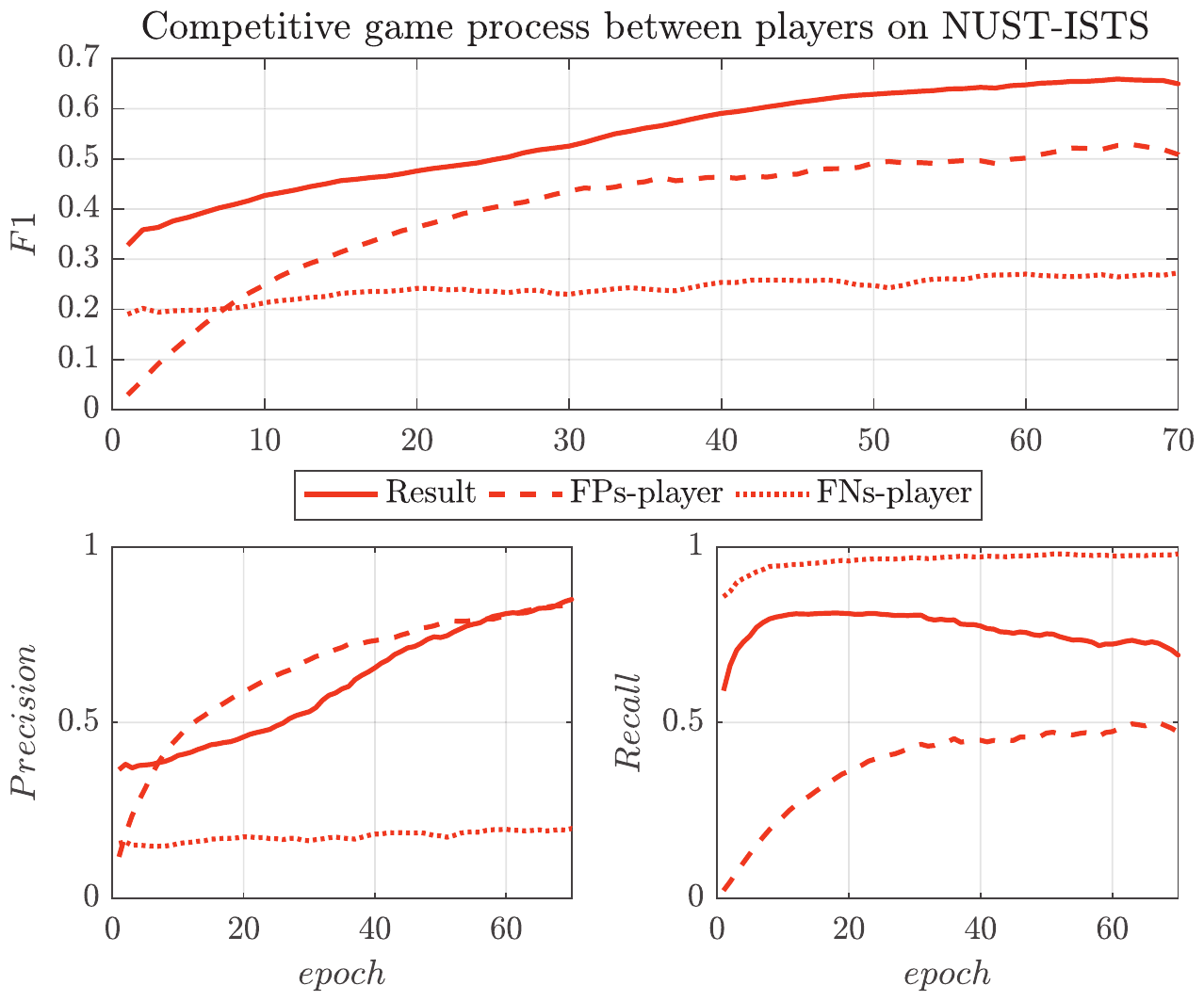}} 
		\subfigure[NUAA-ISTS]
		{\includegraphics[scale=0.65]{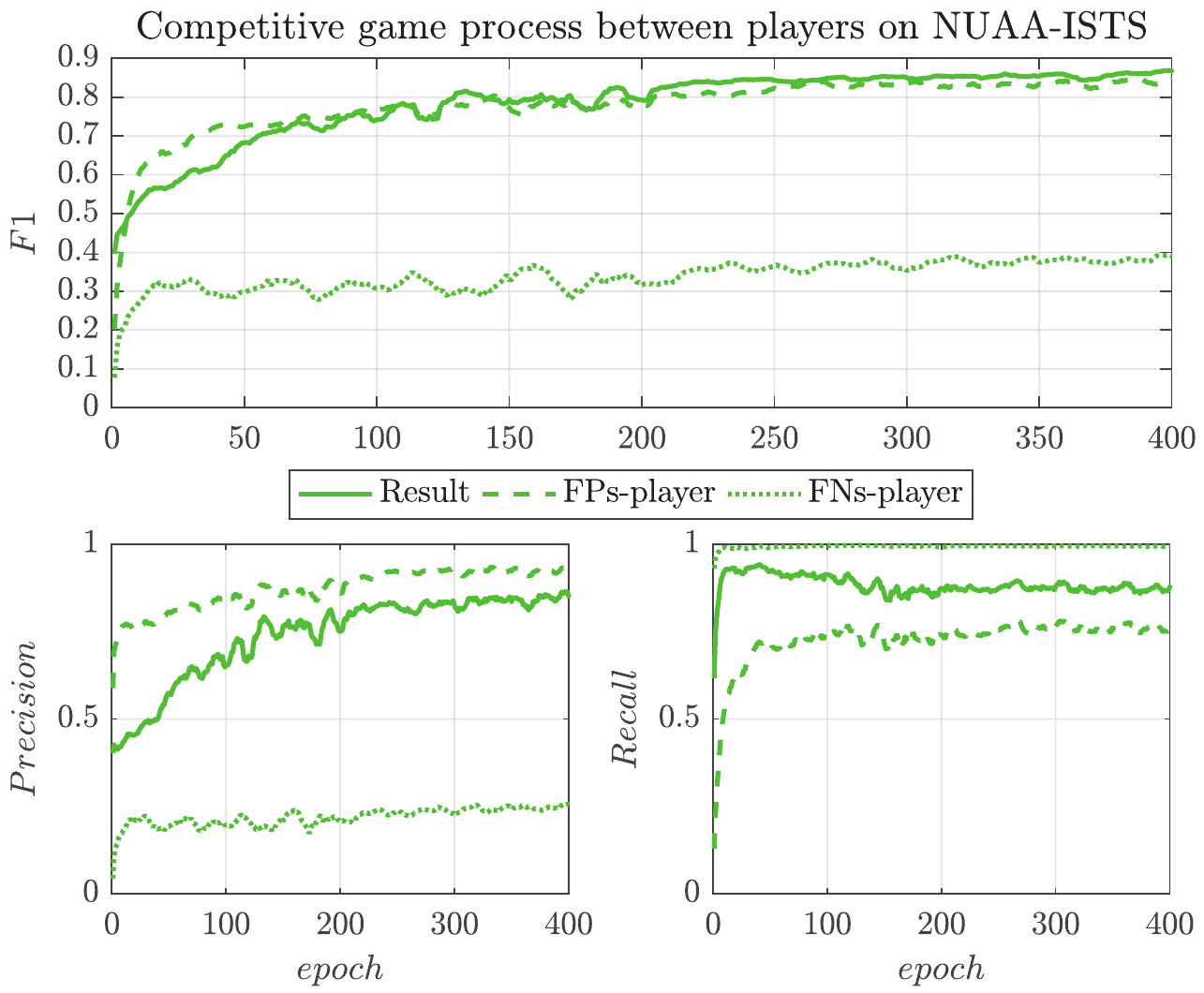}}
		\caption{
			\textbf{Evolution of player performance and Nash equilibrium.}
			(a) and (b) show the competitive game process between players on NUST-ISTS and NUAA-ISTS datasets, respectively.
		} 
		\label{Fig.PRF1} 
	\end{figure*}
	
	\subsubsection{Evaluation Metrics}
	We use precision ($\mathrm{P}$), recall ($\mathrm{R}$), F1 score ($\mathrm{F_1}$) and intersection over union ($\mathrm{IoU}$) to evaluate the infrared small target segmentation methods.
	
	The precision measures the proportion of correctly segmented target pixels in all segmented target pixels. 
	The recall measures the proportion of correctly segmented target pixels in all true target pixels. 
	In Eq. \eqref{Eq.Precision} and Eq. \eqref{Eq.Recall}, $ \mathrm{P} $ and $ \mathrm{R} $ are defined as
	\begin{equation}
		\mathrm{P}=\frac{\sum_{i=1}^{N} o_{i} g_i}{\sum_{i=1}^{N} o_{i}},
		\label{Eq.Precision}
	\end{equation}
	\begin{equation}
		\mathrm{R}=\frac{\sum_{i=1}^{N} o_{i} g_i}{\sum_{i=1}^{N} g_i}.
		\label{Eq.Recall}
	\end{equation}
	
	Beside, $ \mathrm{IoU} $ is also used to measure the coincidence between the predicted mask and the ground truth. $ \mathrm{IoU} $ is defined in Eq. \eqref{Eq.IoU}.
	\begin{equation}
		\mathrm{IoU}= \frac{\sum_{i=1}^{N} o_{i} g_i}{\sum_{i=1}^{N}\left(o_{i}+g_i-o_{i} g_i\right)}.
		\label{Eq.IoU}
	\end{equation}
	In Eq. \eqref{Eq.Precision}, \eqref{Eq.Recall} and \eqref{Eq.IoU}, $o_i$ denotes the class probability of the $i$-th pixel in the segmentation result $ \textit{\textbf{O}} $. 
	$g_i$ is the class probability of the corresponding position of ground truth $ \textbf{\textit{G}} $.
	
	In order to evaluate the advantages and disadvantages of different algorithms, 
	the concept of $\mathrm{F_1}$ value is proposed based on Eq.$\eqref{Eq.Precision}$ and Eq.$\eqref{Eq.Recall}$. $\mathrm{F_1}$ evaluates $ \mathrm{P} $ and $ \mathrm{R} $ together. 
	$\mathrm{F_1}$ is defined as follows:
	\begin{equation}
		\mathrm{F_1}=2 /\left(\frac{1}{\mathrm{P}}+\frac{1}{\mathrm{R}}\right)=2 \frac{\mathrm{P} \times \mathrm{R}}{\mathrm{P}+\mathrm{R}}.
		\label{Eq.F1}
	\end{equation}

	\subsection{Ablation Study}
	In this section, we study questions Q1-Q3 raised above.
	\subsubsection{Players Game and Nash Equilibrium (Q1)}
	\label{section.NashEq}

	In Fig. \ref{Fig.PRF1} (a) and (b), there is a clear trend of gradual steady state.
	As the game between players goes deep, the performance of the players is improved gradually, and finally reaches the Nash equilibrium. 
	From a single index, FNs-player has the highest precision and FPs-player has the highest recall.
	The game result is between FNs-player and FPs-player. 
	But from the final $\mathrm{F_1}$ score, we can see that the results of pixelGame has achieved better performance than the two components as a whole.
	$\mathrm{F_1}$ score in Fig.~\ref{Fig.PRF1} (a) and (b) illustrates the feasibility and superiority of our separate optimization method. 
	
	For different players, FNs-player tends to predict pixels of all potential targets, 
	so it can be seen from recall in Fig.~\ref{Fig.PRF1} (a) and (b) that more than 98\% of the true target pixels are correctly predicted by FNs-player. 
	Differently, FPs-players have high precision and low recall scores. FPs-players put quality before quantity.
	Therefore, the precision of FPs-player on two datasets is more than 85\%.
	Two players with opposing strategies promote pixelGame to achieve a better balance in the overall game framework.
	
	\begin{figure}[t]  
		\centering 
		\subfigure[NUST-ISTS]
		{\includegraphics[scale=0.292]{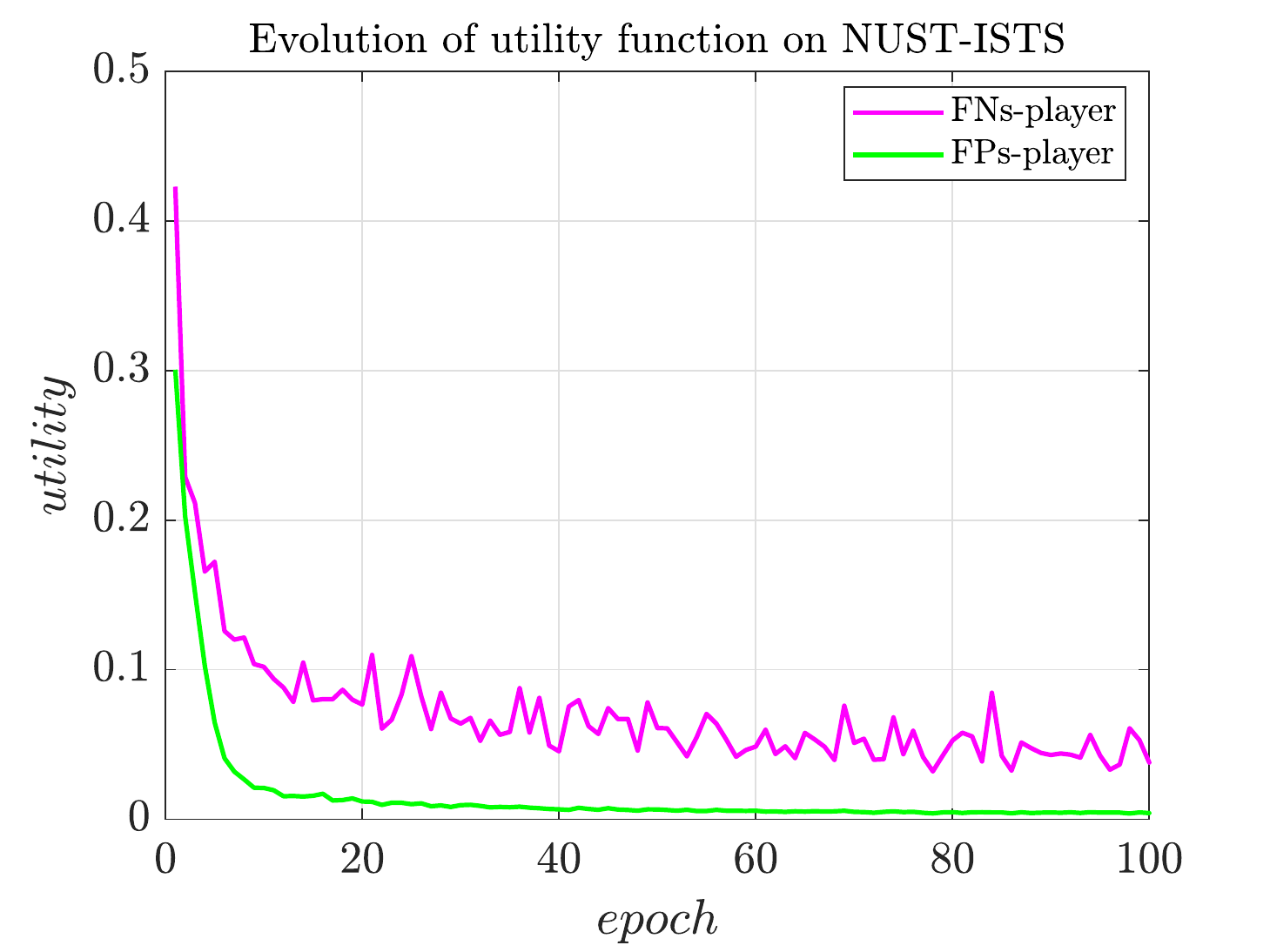}}  
		\subfigure[NUAA-ISTS]
		{\includegraphics[scale=0.292]{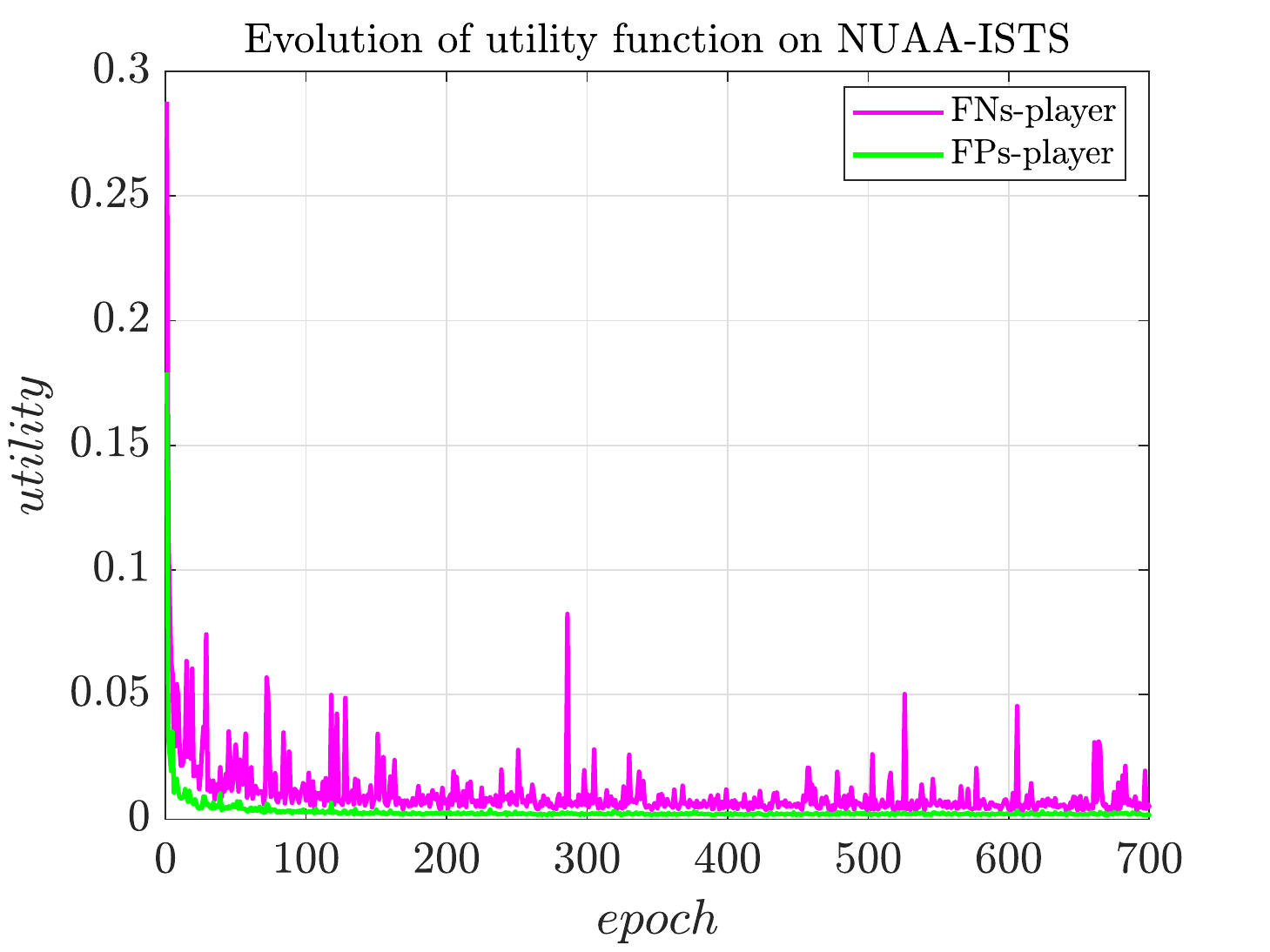}} 
				\vspace{-0.3cm} 
		\caption{{
				Convergence of the utility function in pixelGame.
			}
		} 
		\label{Fig.Loss} 
	\end{figure}

	Furthermore, from the score of $\mathrm{F_1}$ in the Fig. \ref{Fig.PRF1}, 
	the antagonistic learning of players at the pixel level makes the two players have better complementarity at the image level. 
	The fusion result of the two players is considerably better than that of a single player. 
	As show in Fig. \ref{Fig.PRF1} (a), the fusion results of FNs-player and FPs-player are improved by about 0.14 and 0.37 on the NUST-ISTS dataset, respectively.
	The Fig. \ref{Fig.PRF1} (b) presents the fusion results of two players are improved by about 0.05 and 0.46 on the NUAA-ISTS dataset, respectively.
	The performance improvement proves the effectiveness of multi-player adversarial learning based on game theory. 
	Under the guidance of the proposed utility function, 
	the model achieves higher accuracy and a more delicate balance between $ FNs $ and $ FPs $.
	
	{
		Nash equilibrium \cite{kallel2014nash}, \cite{hsieh2021adaptive} represents a static state, where no player is willing to change strategy any more.
		During the pixelGame, Fig. \ref{Fig.Loss} (a) and (b) illustrate the evolution of player utilities for FNs-player and FPs-player on the NUST-ISTS and NUAA-ISTS datasets, respectively. 
		As shown in Fig. \ref{Fig.Loss} (a) and (b), the utilities of the two players in the game tend to be steady after decline, and finally reach a stable state.
		The invariance of the utility functions of two players means that the players do not want to change their strategies, which means the pixelGame reaches a Nash equilibrium.
	}
	
	\subsubsection{Impact of Utility Function (Q2)}
	\label{section.Q2}
		
	In order to understand which utility function is critical to the game, 
	we analyze the results on the NUST-ISTS and NUAA-ISTS datasets for each component of the utility function.
	
\begin{figure*}[t]  
		\centering 
		\subfigure[NUST-ISTS]
		{\includegraphics[width=8.5cm]{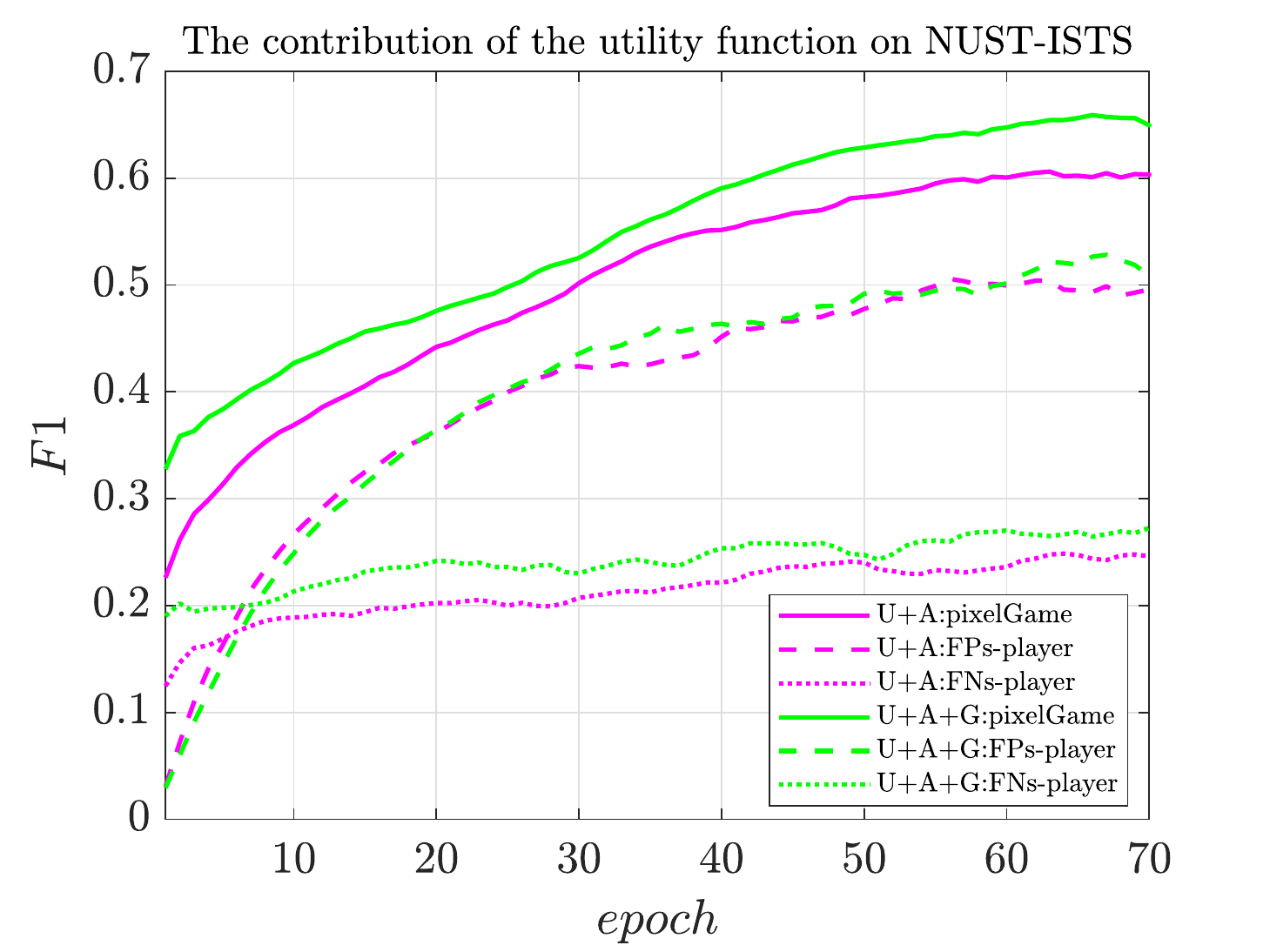}}
		\subfigure[NUAA-ISTS]
		{\includegraphics[width=8.5cm]{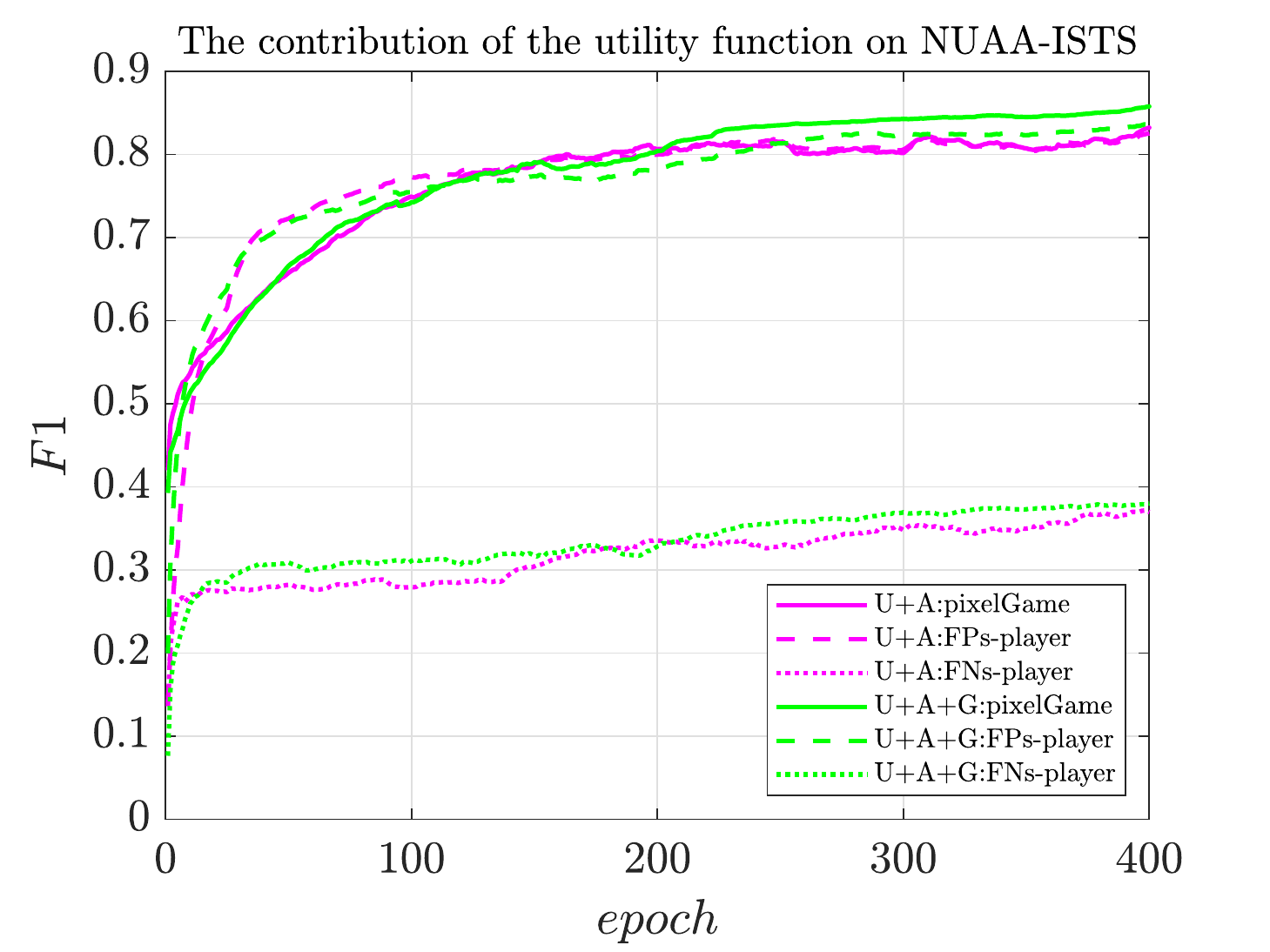}}
		\caption{\textbf{The contribution of game utility to model performance.}
			$U$, $A$ and $G$ represent player utility, small target constraint and game utility respectively.
		}
		\label{Fig.utility}
	\end{figure*}

	The contribution of different components of utility function to the model is shown in TABLE \ref{table.utility}. 
	It can be seen from the TABLE \ref{table.utility} that competitive games under play utility constraints have achieved high results on $\mathrm{F_1}$ and $\mathrm{IoU}$. 
	$ FNs $ and $ FPs $, as two kinds of incorrectly segmented pixels, are the two opposing sides in the overall goal of the game.
	As shown in TABLE \ref{table.utility}, under the guidance of player utility, most of the incorrectly segmented pixels are correctly segmented. 
	It can be observed from TABLE~\ref{table.utility} and Fig.~\ref{Fig.utility} that game utility makes the results of the two more complementary, and increases by about 5\% and 10\% on NUST-ISTS and NUAA-ISTS, respectively.
	The game utility makes the direct competition between the two players more purposeful.
	The area constraint of the small target is conducive to the faster convergence of the player network, 
	and it improves the performance of the model to a certain extent.
	
	We further analyze the impact of game utility on FNs-player and FPs-player.
	The game utility urges the two players to pay attention to different predicted pixels. 
	As shown in Fig.~\ref{Fig.utility} (a) and (b), 
	the game utility results in a better Nash equilibrium state for pixelGame.
	
	\begin{table}[t]
		\centering
		\caption{
			Results of ablation experiments for utility functions.
			In the table, $U$, $G$ and $A$ represent player utility, game utility and small target constraint, respectively.
			Bold font highlights the best results in each column. 
		}
		\label{table.utility}
		\setlength{\tabcolsep}{4mm}{%
			\begin{tabular}{lcccc}
				\toprule
				\multirow{2}{*}{Utility functions} & \multicolumn{2}{c}{NUST-ISTS} & \multicolumn{2}{c}{NUAA-ISTS} \\ \cmidrule(lr){2-3} 		\cmidrule(lr){4-5} 
				\multicolumn{1}{c}{}	& $\mathrm{F_1}$			& $\mathrm{IoU}$		& $\mathrm{F_1}$		& $\mathrm{IoU}$         \\ \midrule
				$U$						& 0.5618					& 0.4531				& 0.7418				& 0.6129        \\
				$U+A$					& 0.6047					& 0.4796				& 0.8095				& 0.7400        \\
				$U+G$					& 0.6272					& 0.5092				& 0.8493				& 0.7339        \\
				$U+A+G$					& \textbf{0.6387}			& \textbf{0.5135}		& \textbf{0.8659}		& \textbf{0.7452}        \\ \bottomrule
		\end{tabular}}
	\end{table}	
	
	
	\subsubsection{The Advantage of Separate Game Objectives Compared to Combined Objectives (Q3)}
	\label{section.Q3}
	Next, we compare the performance of different classical combined objective functions, including
	$Dice \ loss$ \cite{isensee2021nnu}, $IoU \ loss$ \cite{rahman2016optimizing}, 
	and sensitivity specificity loss ($SS \ loss$) \cite{brosch2015deep}, etc.
	The parameter controlling the balance between sensitivity and specificity in $SS \ loss$ is set to $0.5$.
	
	TABLE \ref{table.loss_functoin} shows that the model performance is poor under the guidance of the combined loss functions. 
	In ISTS, infrared targets are dim, small and sparse within images. These challenges make commonly used combined loss functions unable to effectively focus a small number of target pixels, and cannot distinguish background and noise.
	
	\begin{table}[t]
		\caption{{
			Results $(\mathrm{F_1}/\mathrm{IoU})$ of ablation experiments for combined objectives ( $IoU \ loss$, $Dice \ loss$ and $SS \ loss$ ) and separated game objective ( $ Ours $). 
			The best results are highlighted in boldface.
			}
		}
		\label{table.loss_functoin}
		\centering
		\setlength{\tabcolsep}{2.5mm}{
			\begin{tabular}{ccccc}
				\toprule
				\multirow{2}{*}{Objectives} 	& \multicolumn{2}{c}{NUST-ISTS} & \multicolumn{2}{c}{NUAA-ISTS} \\ \cmidrule(lr){2-3} \cmidrule(lr){4-5} 
				& FNs-player    				& FPs-player    & FNs-player	& FPs-player    \\ \midrule
				$IoU \ loss$                    & 0.54/0.45						& 0.50/0.41		& \textbf{0.77}/\textbf{0.66}		& 0.76/0.65     \\
				$Dice \ loss$                   & \textbf{0.55}/\textbf{0.46}	& 0.48/0.40		& 0.76/0.65		& 0.76/0.66     \\ 
				$SS \ loss$                     & 0.33/0.22						& 0.32/0.21		& 0.37/0.25		& 0.35/0.23     \\
				{$Ours$}   	& {0.27/0.17}& {\textbf{0.51}/\textbf{0.43}}		& {0.38/0.29}		& {\textbf{0.82}/\textbf{0.72}}     \\ \bottomrule
		\end{tabular}}
	\end{table}
	
	$ \mathrm{IoU} $ and $ \mathrm{Dice} $ similarity coefficient look very similar in terms of equations, and both are the most commonly used evaluation metrics in object segmentation.
	As reported in TABLE \ref{table.loss_functoin}, using $IoU \ loss$ and $Dice \ loss$, in NUST-ISTS and NUAA-ISTS datasets, FNs-player and FPs-player only reach about 0.50 and 0.76 on $\mathrm{F_1}$,
	and the $\mathrm{IoU}$ of these two sub-networks are about 0.45 and 0.65, respectively.
	The optimization objectives of both mainly focus on $ TPs $, treating two classes of incorrectly segmented pixels equally.
	However, due to the IR targets are small, the model leads to over-segmentation easily. The pixels of over-segmentation are mainly noise and background near the target, which are mainly reflected in $ FPs $. 
	The model cannot effectively balance a small number of but very critical pixels such as the edges of dim small targets ($ FPs $ and $ FNs $), resulting in stagnant performance.
	
	Although $ FNs $ and $ FPs $ are considered in $SS \ loss$ explicitly, their performance is not ideal.
	From the perspective of features, features of false positive and false negative pixels are similar, but they are different or even opposite in optimization strategies. 
	For ISTS, the $SS \ loss$ function uses the same strategy to optimize the countermeasure decision, so that the model cannot converge to the global optimal solution.
	
	Comparing the results of combined objectives and separate optimization in TABLE \ref{table.loss_functoin},
	it can be seen that FNs-player achieves leading performance by using combined objectives.
	Considering the beneficial balance of FNs-payer to FPs-player, the performance of pixelGame is further improved.
	This also proves the prominent performance of our FNs-player and FPs-player network structure for ISTS. 
	\begin{table*}[t]
		\centering
		\caption{Comparison with the state-of-the-art ISTS methods.
			The best result in each column is in {\color{red} \textbf{red}}, 
			the second is in {\color{blue} blue}, 
			and The third is in {\color{green} green}.
			{
			The pixelGame (w/o $ \mathrm{MIM} $) is the baseline model.}
		}
		\label{table.SOAT}
		\setlength{\tabcolsep}{4mm}{%
			\begin{tabular}{r|cccc|cccc}
				\toprule
				\multicolumn{1}{c|}{}                         & \multicolumn{4}{c|}{NUST-ISTS}                                                                                                                            & \multicolumn{4}{c}{NUAA-ISTS}                                                                                                                             \\ 
				\multicolumn{1}{c|}{\multirow{-2}{*}{Method}} & Precision                            & Recall                               & $ \mathrm{F_1} $                     & IoU                       & Precision                            & Recall                               & $ \mathrm{F_1} $                           & IoU                      \\ \midrule
				Top-Hat\cite{rivest1996detection} (OE 1996)   & 0.09                                 & 0.21                                 & 0.12                                 & 0.08                                 & 0.14                                 & 0.11                                 & 0.12                                 & 0.13                                 \\
				Max-Median\cite{deshpande1999max} (ISOP 1999) & 0.05                                 & 0.14                                 & 0.05                                 & 0.03                                 & 0.04                                 & 0.18                                 & 0.04                                 & 0.03                                 \\
				MNWTH\cite{bai2010analysis} (PR 2010)         & 0.23                                 & 0.61                                 & 0.27                                 & 0.18                                 & 0.18                                 & 0.27                                 & 0.22                                 & 0.27                                 \\
				IPI\cite{gao2013infrared} (TIP 2013)          & 0.51                                 & 0.49                                 & 0.50                                 & 0.38                                 & 0.31                                 & 0.48                                 & 0.34                                 & 0.28                                 \\
				LCM\cite{chen2013local} (TGRS 2013)           & 0.15                                 & 0.36                                 & 0.21                                 & 0.13                                 & 0.15                                 & 0.29                                 & 0.22                                 & 0.14                                 \\
				ILCM\cite{han2014robust} (GRSL 2014)          & 0.14                                 & 0.22                                 & 0.20                                 & 0.13                                 & 0.14                                 & 0.25                                 & 0.21                                 & 0.13                                 \\
				MPCM\cite{wei2016multiscale} (PR 2016)        & 0.28                                 & 0.45                                 & 0.34                                 & 0.24                                 & 0.29                                 & 0.37                                 & 0.34                                 & 0.27                                 \\
				NWIE\cite{deng2016infrared} (TAES 2016)       & 0.23                                 & 0.21                                 & 0.21                                 & 0.18                                 & 0.31                                 & 0.21                                 & 0.23                                 & 0.14                                 \\
				NIPPS\cite{dai2017non} (INFPHY 2017)          & 0.12                                 & 0.29                                 & 0.15                                 & 0.10                                 & 0.29                                 & 0.31                                 & 0.32                                 & 0.21                                 \\
				TV-PCP\cite{wang2017pcp} (ICV 2017)           & 0.38                                 & 0.25                                 & 0.34                                 & 0.22                                 & 0.32                                 & 0.47                                 & 0.38                                 & 0.40                                 \\
				SMSL\cite{wang2017infrared} (TGRS 2017)       & 0.54                                 & 0.20                                 & 0.26                                 & 0.17                                 & 0.40                                 & 0.45                                 & 0.41                                 & 0.40                                 \\
				RLCM\cite{han2018infrared} (GRSL 2018)        & 0.39                                 & 0.45                                 & 0.41                                 & 0.31                                 & 0.46                                 & 0.44                                 & 0.44                                 & 0.31                                 \\
				MoG-MRF\cite{gao2018infrared} (PR 2018)       & 0.22                                 & 0.45                                 & 0.28                                 & 0.20                                 & 0.25                                 & 0.34                                 & 0.31                                 & 0.23                                 \\
				DPIR\cite{huang2019infrared} (GRSL 2019)      & 0.52                                 & 0.21                                 & 0.24                                 & 0.17                                 & 0.45                                 & 0.32                                 & 0.37                                 & 0.27                                 \\
				TLLCM\cite{han2019local} (GRSL 2019)          & 0.52                                 & 0.42                                 & 0.41                                 & 0.35                                 & 0.60                                 & 0.62                                 & 0.54                                 & 0.44                                 \\
				WSLCM\cite{han2020infrared} (GRSL 2020)       & 0.38                                 & 0.30                                 & 0.33                                 & 0.22                                 & 0.58                                 & 0.42                                 & 0.50                                 & 0.38                                 \\
				WSNMSTIPT\cite{sun2019infrared} (INFPHY 2019) & 0.28                                 & 0.15                                 & 0.15                                 & 0.13                                 & 0.26                                 & 0.25                                 & 0.21                                 & 0.25                                 \\
				MDvsFA-cGAN\cite{wang2019miss} (ICCV 2019)    & {\color[HTML]{3531FF} 0.63}          & 0.65                                 & {\color[HTML]{3531FF} 0.61}          & {\color[HTML]{34FF34} 0.47}          & 0.72                                 & 0.77                                 & 0.71                                 & 0.60                                 \\
				MDWCM\cite{lu2020robust} (GRSL 2020)          & 0.38                                 & 0.42                                 & 0.40                                 & 0.43                                 & 0.32                                 & 0.43                                 & 0.37                                 & 0.41                                 \\
				ECA-STT\cite{zhang2020edge} (TGRS 2020)       & 0.53                                 & 0.46                                 & 0.49                                 & 0.38                                 & 0.79                                 & 0.58                                 & 0.64                                 & 0.53                                 \\
				ADMD\cite{moradi2020fast} (SP 2020)           & 0.57                                 & 0.22                                 & 0.27                                 & 0.19                                 & 0.69                                 & 0.38                                 & 0.42                                 & 0.30                                 \\
				ACM\cite{dai2021asymmetric} (WACV 2021)       & 0.55                                 & {\color[HTML]{FE0000} \textbf{0.74}} & {\color[HTML]{34FF34} 0.60}          & 0.44                                 & {\color[HTML]{34FF34} 0.83}          & {\color[HTML]{34FF34} 0.84}          & {\color[HTML]{34FF34} 0.83}          & 0.71                                 \\
				SRWS\cite{zhang2021infrared} (NC 2021)        & 0.56                                 & 0.34                                 & 0.44                                 & 0.33                                 & 0.74                                 & 0.47                                 & 0.53                                 & 0.39                                 \\
				NTFRA\cite{kong2021infrared} (TGRS 2021)      & 0.47                                 & 0.43                                 & 0.46                                 & 0.38                                 & 0.67                                 & 0.55                                 & 0.57                                 & 0.44                                 \\
				ALCNet\cite{dai2021attentional} (TGRS 2021)   & 0.56                                 & {\color[HTML]{3531FF} 0.73}          & {\color[HTML]{34FF34} 0.60}          & 0.45                                 & {\color[HTML]{FE0000} \textbf{0.88}} & {\color[HTML]{34FF34} 0.84}          & {\color[HTML]{3531FF} 0.85}          & {\color{green} 0.73}          \\ \midrule
				\multicolumn{1}{c|}{pixelGame (w/o $ \mathrm{MIM} $)}                                      & {\color[HTML]{34FF34} 0.60}          & {\color[HTML]{34FF34} 0.72}          & {\color[HTML]{3531FF} 0.61}          & {\color[HTML]{3166FF} 0.48}          & {\color[HTML]{3166FF} 0.86}          & {\color[HTML]{3531FF} 0.85}          & {\color[HTML]{3531FF} 0.85}          & {\color[HTML]{3531FF} 0.74}          \\
				\multicolumn{1}{c|}{pixelGame}                      & {\color[HTML]{FE0000} \textbf{0.66}} & {\color[HTML]{FE0000} \textbf{0.74}} & {\color[HTML]{FE0000} \textbf{0.64}} & {\color[HTML]{FE0000} \textbf{0.51}} & {\color[HTML]{FE0000} \textbf{0.88}} & {\color[HTML]{FE0000} \textbf{0.86}} & {\color[HTML]{FE0000} \textbf{0.86}} & {\color[HTML]{FE0000} \textbf{0.75}} \\ \bottomrule
			\end{tabular}%
		}
	\end{table*}
		
	\subsection{Comparison to State-of-the-Art Approaches}
	\label{section.SOAT}
	Finally, we solve the problem Q4 by comparing our pixelGame with several CNN-based methods and other traditional mathematical modeling methods.
	For a long time, the lack of an open benchmark has been one of the bottlenecks hindering the development of ISTS, 
	which allows various algorithms to be compared fairly. 
	We have summarized the existing target detection and segmentation methods of infrared small targets, 
	which is helpful to promote ISTS work. 
	The comparison between the pixelGame and the SOTA methods is reported in TABLE \ref{table.SOAT}. 
	In experiments, we evaluate the performance of the algorithm at the pixel level using precision, recall, $\mathrm{F_1}$ and $\mathrm{IoU}$.
	
	Compared with traditional algorithms, 
	pixelGame is comfortably ahead in both single score (precision and recall) and comprehensive score ($\mathrm{F_1}$ and $\mathrm{IoU}$). 
	In CNN-based models, pixelGame better suppresses $ FNs $ and $ FPs $, 
	and achieves a more delicate balance between precision and recall.
	In TABLE~\ref{table.SOAT}, on both the NUST-ISTS and NUAA-ISTS datasets, our method achieves the highest $\mathrm{F_1}$ and $\mathrm{IoU}$ on both datasets, which are 0.64, 0.86 and 0.51, 0.75, respectively. 
	For the foreground-background imbalance problem, our pixelGame achieves the best balance between $ FNs $ and $ FPs $. 
	From the comparison of precision and recall, which are 0.66, 0.74 and 0.88, 0.86, we can see that the precision and recall of our model are closer and more balanced.
	
	Moreover, it can be seen from TABLE \ref{table.SOAT} that deep learning models perform better than the traditional algorithms. Specifically, 
	the CNN-based algorithms mostly design the loss function of the model at the pixel level, 
	such as MDvsFA-cGAN \cite{wang2019miss} (based on miss detection and false alarm), ACM \cite{dai2021asymmetric} and ALCNet \cite{dai2021attentional} (based on $nIoU$). 
	This demonstrates that the pixel-wise loss function has significant advantages in dense segmentation tasks. 
	The traditional algorithms, such as Top-Hat \cite{rivest1996detection} and LCM~\cite{chen2013local}, mostly 
	suppress the background, and enhance the difference between the target and the background at the image level. 
	Similar to the algorithm of recovering the target from the feature space, 
	such as IPI~\cite{gao2013infrared} and IPT~\cite{zhang2021infrared} series, 
	the target position and approximate shape can be determined, but the precise control at the pixel level cannot be achieved. 
	
	In the CNN-based methods, most of the existing methods 
	treat different kinds of error pixels as the same to optimize. 
	For example, ACM \cite{dai2021asymmetric} and ALCNet \cite{dai2021attentional} use $IoU \ loss$ as the optimization function. 
	The integrated loss methods are not suitable for conquering extreme foreground-background imbalance problem of ISTS.
	MDvsFA-cGAN \cite{wang2019miss} makes the two sub-networks focus on false positive pixels and false negative pixels by loss re-weighting, respectively. 
	However, in essence, MDvsFA-cGAN optimizes $ FNs $ and $ FPs $ at the same time, which is not pure antagonistic learning between $ FNs $ and $ FPs $. 
	Differently,
	our method based on game theory optimizes false positive pixels and false negative pixels separately. 
	The pixelGame is conducive to flexible selection of different optimization strategies and is easier to achieve Nash equilibrium.
	
	\begin{figure*}[b]  
		\centering 
		\includegraphics[scale=0.28]{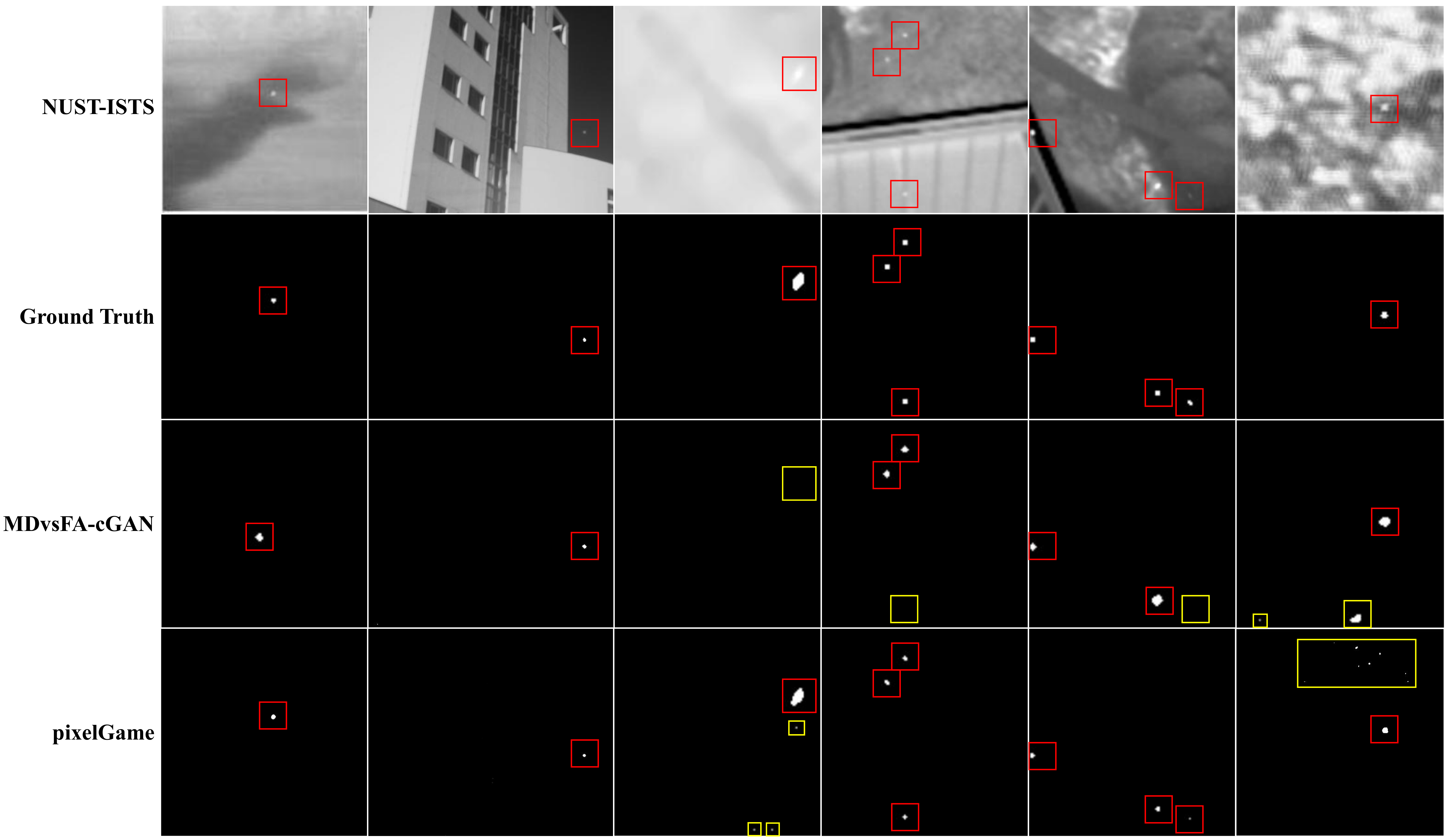}  
		\caption{\textbf{Segmentation results on NUST-ISTS.}
			The infrared targets in the images are marked by a red box.
			The yellow box represents the wrong segmentation result.
		} 
		\label{Fig.res_images1} 
	\end{figure*}
	
	{
		To demonstrate the effectiveness of $ \mathrm{MIM} $, we visualize the feature maps enhanced by $ \mathrm{MIM} $ module.
		It can be observed from Fig. \ref{Fig.Vis_MIM} that $ \mathrm{MIM} $ performs better in capturing IR small targets with low SCR than other attention methods.
		The main content in infrared image is the building, while the region of target is very small compared with the main building.
		SENet, GCNet and Triplet attention tend to focus on the main building and neglect the small target.
		Differently, $ \mathrm{MIM} $ prominently enhances the target information and suppresses the background.
		Benefiting from the different dilation factors, $ \mathrm{MIM} $ effectively focuses on the salient region including small targets, which better suits our purpose.
		Fig. \ref{Fig.Vis_MIM} suggests that the enhanced features by $ \mathrm{MIM} $ are powerful.
	}

	{
	In addition, we design ablation experiments by removing $ \mathrm{MIM} $ (denoted as w/o $ \mathrm{MIM} $).
	We apply simple addition when removing each module. 
	From the last two rows of TABLE~\ref{table.SOAT}, 
	compared to pixelGame (w/o $ \mathrm{MIM} $), pixelGame improves the segmentation accuracy by about 3\% and 1\% respectively on the two datasets.
	As can be seen from TABLE~\ref{tab:att}, $ \mathrm{MIM} $ achieves the best performance on both datasets. The F1 score of MIM is about 2\% higher on average than other methods.
	TABLE~\ref{table.SOAT} and TABLE \ref{tab:att} present the effectiveness of $ \mathrm{MIM} $ for local significant context information enhancement in ISTS tasks.
	The efficient of $ \mathrm{MIM} $ is fully proved.
}	

	\begin{table}[t]
		\centering
		\caption{{
			Results of ablation experiments for attention modules.
			Bold font highlights the best results in each column. 
		}}
		\label{tab:att}
		\setlength{\tabcolsep}{4.8mm}{%
			{\begin{tabular}{lcccc}
					\toprule
					\multicolumn{1}{c}{\multirow{2}{*}{Method}} & \multicolumn{2}{c}{NUST-ISTS} & \multicolumn{2}{c}{NUAA-ISTS} \\ \cmidrule(lr){2-3} \cmidrule(lr){4-5}
					\multicolumn{1}{c}{}                        & $ \mathrm{F_1} $            	& IoU           & $ \mathrm{F_1} $            	& IoU           \\ \midrule
					w/ SENet                                    & 0.59          		& 0.50          & 0.81          		& 0.73          \\
					w/ GCNet                                    & 0.62          		& 0.51          & 0.83          		& 0.74          \\
					w/ Triplet                                  & 0.55          		& 0.47          & 0.75          		& 0.67          \\
					w/ MIM                                      & \textbf{0.64} 		& \textbf{0.51} & \textbf{0.86} 		& \textbf{0.75}          \\ \bottomrule
		\end{tabular}}}
	\end{table}
	\begin{table}[t]
		\centering
		\caption{{
				Complexity analysis of CNN-based models on NUST-ISTS datasets.
				Params refer to the total number of parameters that need to be trained in model training.
				The speed (FPS) of the model is measured on RTX 2080 Ti.
				FLOPs are floating point operations.
				The best results are highlighted in boldface.
			}}
		\label{table:time}
		\setlength{\tabcolsep}{3mm}{%
			{\begin{tabular}{ccccc}
					\toprule
					Method       & Params 			& FPS   			& FLOPs   			& IoU			\\ \midrule
					ACM          & 0.63M  			& \textbf{100}    	& 1.75G   			& 0.44			\\
					ALCNet       & \textbf{0.52M}  	& \textbf{100}    	& \textbf{1.41G}   	& 0.45			\\
					MDvsFA-cGAN  & 3.76M  			& 2.63    			& 868.75G 			& 0.47			\\
					pixelGame    & 1.69M  			& 9.09    			& 273.15G 			& \textbf{0.51}		\\ \bottomrule
		\end{tabular}}}
	\end{table}
	{
	We further study the spatiotemporal complexity of CNN-based methods. The results are shown in TABLE \ref{table:time}.
	ACM and ALCNet use ResNet-20 \cite{he2016deep} as the backbone architecture.
	Both MDvsFA-cGAN and pixelGame employ dilated convolutional networks.
	MDvsFA-cGAN needs more parameters due to more channels of features and fully connected layers in the discriminator. 
	The computational complexity is relatively high. 
	It can be seen from TABLE \ref{table:time} that pixelGame trades off speed for a high improvement in segmentation accuracy.
}
	
\begin{figure*}[t]  
		\centering 
		\includegraphics[scale=0.28]{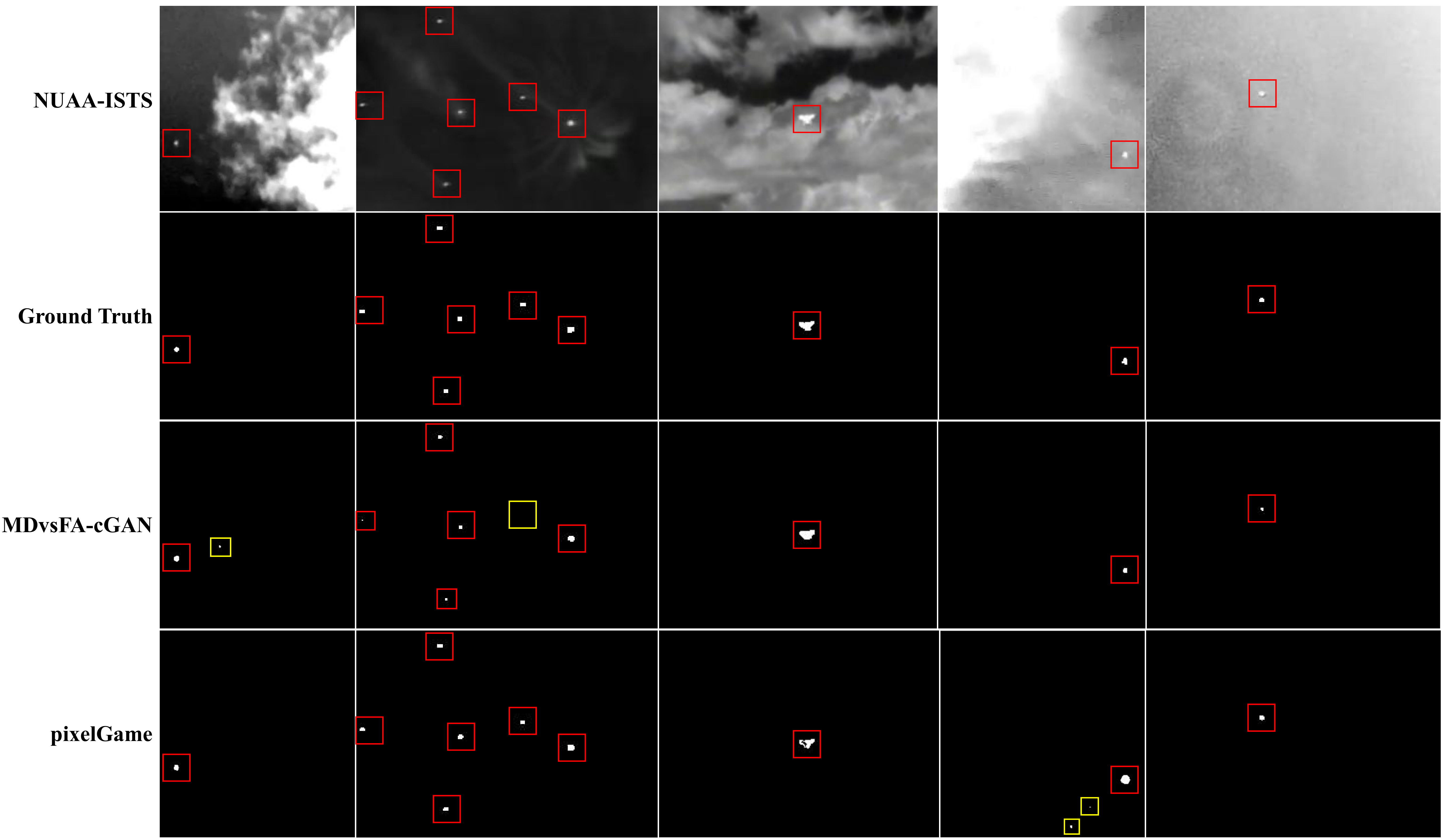}  
		\caption{\textbf{Segmentation results on NUAA-ISTS.}
			The infrared targets in the images are marked by a red box.
			The yellow box represents the wrong segmentation result.
		} 
		\label{Fig.res_images2} 
	\end{figure*}

	Some segmentation results are shown in Fig. \ref{Fig.res_images1} and Fig.~\ref{Fig.res_images2}. 
	It can be illustrated from the Fig. \ref{Fig.res_images1} and Fig. \ref{Fig.res_images2} that the pixelGame segments dim small targets with low SCR accurately and robustly.
	Due to the low spatial resolution, the targets of NUST-ISTS dataset are small and dim.
	Some bright background noise is incorrectly predicted as target pixels. 
	The background of NUAA-ISTS dataset mostly comes from real scenes, 
	the contrast between the target and the background is relatively larger, and the segmentation results are more refined.
	
	By and large, the method based on deep learning has obvious advantages in feature representation compared with the traditional methods. 
	Combining the advantages of deep learning and attention model,
	our method based on game theory achieves a better balance in precision and recall, and achieves the best performance in $\mathrm{F_1}$ and $\mathrm{IoU}$. 
	For small targets in infrared images with low SCR,
	our proposed $ \mathrm{MIM} $ effectively improves the quality of extracted features.

	\section{Conclusion}
	\label{section.Conclusion}
	In ISTS, the IR targets are small in size, weak in energy, and sparse in layout. 
	To solve those problems, we formulate the infrared small target segmentation into playing a competitive game problem.
	we transform infrared small target segmentation into a competitive game problem. 
	Under the guidance of utility function, each player optimizes different pixels. 
	In the continuous game confrontation, the network continues to learn and finally reaches Nash equilibrium. 
	The proposed $ \mathrm{MIM} $ has good performance in sensing small target signals.
	Compared with traditional methods and deep learning methods based on combined loss function, our pixelGame achieves a better balance in precision and recall, and the highest $\mathrm{F_1}$ and $\mathrm{IoU}$. 
	
	For future work, we will try to further improve the efficiency of game optimization, 
	and introduce more prior knowledge to make up the lacking of infrared small target information.

	\section*{Acknowledgment}
	The authors would like to thank the editor and the anonymous reviewers 
	for their constructive comments, which are very helpful to improve the quality of this article.
	\bibliographystyle{IEEEtran}
	
	\bibliography{pixelGame}

\end{document}